%% file: pessimismdro.tex
\documentclass[11pt]{article}

\usepackage[margin=1in]{geometry}
\usepackage{times}
\usepackage[authoryear,round]{natbib}
\usepackage{amsmath,amssymb,amsfonts,amsthm,mathtools}
\usepackage{algorithm}
\usepackage{algorithmic}
\usepackage{url}
\usepackage{array}
\usepackage{xcolor}
\usepackage{hyperref}
\hypersetup{
  colorlinks=true,
  linkcolor=blue!80!black,
  citecolor=black,
  urlcolor=blue!80!black
}

\newtheorem{assumption}{Assumption}
\newtheorem{definition}{Definition}
\newtheorem{lemma}{Lemma}
\newtheorem{proposition}{Proposition}
\newtheorem{theorem}{Theorem}

\newtheorem{remark}{Remark}

\newcommand{\E}{\mathbb E}
\newcommand{\Pbb}{\mathbb P}
\newcommand{\R}{\mathbb R}
\newcommand{\1}{\mathbf 1}
\newcommand{\cX}{\mathcal X}
\newcommand{\cA}{\mathcal A}
\newcommand{\cY}{\mathcal Y}
\newcommand{\cD}{\mathcal D}

\newcommand{\cU}{\mathcal U}
\newcommand{\cT}{\mathcal T}
\newcommand{\argmax}{\operatorname*{arg\,max}}
\newcommand{\argmin}{\operatorname*{arg\,min}}

\title{Minimax Optimality for Factored Optimal-Transport Distributionally-Robust Policy Learning in Contextual Bandits}
\author{Khang Luong\textsuperscript{1}\quad Trung Nguyen\textsuperscript{1}\quad Hoang Ta\textsuperscript{2}\quad Tuan Quang Dam\textsuperscript{2}\quad Hung The Tran\textsuperscript{1}\\[0.5em]
  \small\textsuperscript{1}Quantum AI \& Cyber Security Institute, FPT Corporation\quad
  \textsuperscript{2}Hanoi University of Science and Technology}
\date{}

\begin{document}
\maketitle

\begin{abstract}
Offline contextual-bandit policies may be deployed under simultaneous shifts in the context distribution and reward laws. We study policy learning for a factored distributionally robust optimization (DRO) objective that models these shifts through separate ambiguity sets. Optimal transport (OT) ambiguity sets are particularly suitable for this purpose, allowing the transport cost to encode application-specific severity of a shift. We develop a pessimistic OT-DRO learner based on sharp confidence bounds for robust rewards and analyze context estimation using a distribution-dependent concentration bound. Requiring knowledge of neither the behavior policy nor the context distribution, our learner achieves a regret of order $D C_{\rm rob}^\star \sqrt{\log(|\cX||\cA|/\delta)/n}$. Our robust optimal-policy coverage coefficient $C_{\rm rob}^\star$ is a natural generalization of the single-policy coverage coefficient in prior non-robust pessimism work, capturing the additional statistical difficulty caused by worst-case context shift. A matching lower bound shows that this rate, including its dependence on $C_{\rm rob}^\star$, is minimax-optimal up to constants. 
\end{abstract}

\section{Introduction}

Offline contextual bandit learning seeks to optimize a decision policy using logged data collected by a possibly different behavior policy \citep{dudik2011doubly,swaminathan2015counterfactual}. It is central to applications in which online experimentation is costly or unsafe, such as recommendation systems, advertising, and personalized treatment \citep{li2010contextual,bottou2013counterfactual,tewari2017ads}. Standard methods typically evaluate policies under the assumption that the deployment environment coincides with the environment that generated the logged data. This assumption is often unrealistic: the population of contexts may change over time, and the reward distribution associated with a given context--action cell may also shift. A policy optimized for the nominal distribution can therefore perform poorly when the context or reward distribution shifts after deployment.

Distributionally robust optimization (DRO) addresses this problem by optimizing the worst-case policy value over a neighborhood of the nominal environment. Early approaches place a single divergence ball around the joint data-generating distribution \citep{si2020distributionally,kallus2022doubly}. Because context shift and conditional reward shift may occur at different scales, factored DRO assigns separate ambiguity sets to the context distribution and the conditional reward laws. Existing factored formulations use Kullback-Leibler (KL) divergence \citep{mu2022factored} or Wasserstein distance \citep{shen2024wasserstein} to formulate the ambiguity sets. We study factored DRO with general optimal-transport (OT) costs, which we call factored OT-DRO. Unlike KL divergence, OT can accommodate distributions with different supports. General transport costs also allow the severity of a shift to be specified more flexibly than the squared-distance cost used by \citet{shen2024wasserstein}. For example, two changes in patient state could have the same squared distance but very different clinical consequences; a chosen transport cost could assign them different costs.

Existing learning guarantees for factored DRO contextual bandits, however, require uniform coverage of all context--action cells \citep{mu2022factored,shen2024wasserstein}. This requirement is restrictive in offline learning because the behavior policy may rarely select actions that are irrelevant to an optimal policy. In the nominal setting, pessimistic policy learning avoids paying for such irrelevant actions by constructing lower confidence bounds and requiring sufficient coverage only along an optimal policy \citep{rashidinejad2021bridging,li2022pessimism,jin2025policy,wang2024oracle}. This suggests that robust policy learning should likewise adapt to the coverage of an optimal robust policy.

Extending this principle to factored OT-DRO is nontrivial. Under context shift, the adversary can place additional mass on contexts that are uncommon in the logged data, so nominal optimal-policy coverage alone does not capture the resulting statistical difficulty. An appropriate coverage condition must account simultaneously for the behavior policy, an optimal robust policy, and every context distribution permitted by the ambiguity set. This raises the central question of this work: \emph{Can factored OT-DRO policies be learned at the minimax-optimal rate without uniform action coverage, paying only for coverage along an optimal robust policy?} We show that the answer is affirmative.

\subsection{Problem Setup}

To state our contributions and compare their guarantees precisely with prior work, we first introduce the logged-data model, the factored OT-DRO objective, and the notation used throughout the paper. We then summarize our contributions and discuss related work. Throughout this paper, let $\Delta(\Omega)$ denote the set of probability distributions over the set $\Omega$. Denote $\cX$ the finite context set, $\cA$ the finite action space, and $\cY = [0,D]$ the reward range with known constant $D<\infty$.

\begin{assumption}[Logged contextual-bandit model]
\label{ass:logged-model}
We observe offline logged bandit data $\cD_n=\{(X_i,A_i,Y_i)\}_{i=1}^n$ consisting of i.i.d. triples generated by
\[
    X_i\sim P^0,
    \qquad
    A_i\mid X_i\sim \pi^0_{X_i},
    \qquad
    Y_i\mid X_i,A_i\sim Q^0_{X_i,A_i},
\]
or equivalently by the joint law $P^0(x)\pi^0_x(a)Q^0_{x,a}(y)$. Here $P^0 \in \Delta(\cX)$ is the nominal context distribution, $\pi^0_x\in \Delta(\cA)$ is the behavior policy conditional on context $x$, and $Q^0_{x,a} \in \Delta(\cY)$ is the nominal reward distribution conditional on context $x$ and action $a$. This data generating model rules out unobserved confounding: conditional on $X_i$, the logging action is drawn from $\pi^0_{X_i}$, and the reward law depends on the sample only through $(X_i,A_i)$.
\end{assumption}

Let $\Pi$ be a nonempty closed class of stochastic policies $\pi:\cX\to\Delta(\cA)$. Our goal is to learn a policy $\pi^\star\in\Pi$ with optimal factored OT-DRO value. The following subsection defines this value.

\subsection{Factored OT-DRO Objective}
\begin{assumption}[Transport costs]
\label{ass:transport-cost}
For $\mathsf S\in\{\cX,\cY\}$, let $c_{\mathsf S}:\mathsf S^2\to[0,\infty)$ be a nonzero lower semicontinuous cost satisfying
\(
c_{\mathsf S}(s,s)=0
\,\text{ and }\,
C_{\mathsf S}:=\sup_{s,t\in\mathsf S}c_{\mathsf S}(s,t)\in(0,\infty).
\)  
\end{assumption}
The transport cost $c_{\mathsf S}$ defines the severity of a shift in the space $\mathsf S$. For example, in a clinical application, $c_{\cX}(x,x')$ may be small if patient states $x$ and $x'$ are clinically similar, while $c_{\cX}(x,x')$ may be large if they are substantially different. Similarly, $c_{\cY}(y,y')$ may be small if rewards $y$ and $y'$ are close in magnitude, while it may be large if they differ substantially.

Let $\mu, \nu \in \Delta(\mathsf S)$ and $\Gamma(\mu,\nu)$ denote the set of all couplings between $\mu$ and $\nu$. Define the transport discrepancy and the corresponding uncertainty set by
\[
\cT_{\mathsf S}(\mu,\nu)
:=
\inf_{\gamma\in\Gamma(\mu,\nu)}
\int c_{\mathsf S}(s,t)\,d\gamma(s,t),
\qquad
\cU_{\mathsf S}(\mu,\epsilon)
:=
\left\{\nu\in\Delta(\mathsf S):
\cT_{\mathsf S}(\mu,\nu)\leq\epsilon\right\}.
\]
Instead of optimizing the nominal policy value $\E_{x\sim P^0}[\E_{a\sim\pi_x}[\E_{y\sim Q^0_{x,a}}[y]]]$, we assume that the deployment distributions of the context and reward lie in the factored OT ambiguity sets $\cU_{\cX}(P^0,\epsilon_{\cX})$ and $\cU_{\cY}(Q^0_{x,a},\epsilon_{\cY})$ for given radii $\epsilon_{\cX}>0$ and $\epsilon_{\cY}>0$. The factored OT-DRO value of a policy $\pi$ is then defined as the worst-case expected reward over these ambiguity sets.
\begin{definition}[Factored OT-DRO value]
\label{def:factored_otdro_value}
With given $\epsilon_{\cX}>0$ and $\epsilon_{\cY}>0$, the factored OT-DRO value of policy $\pi \in \Pi$ is defined as
\[
    V(\pi)
    =
    \inf_{P\in\cU_{\cX}(P^0,\epsilon_{\cX})}
    \E_{x \sim P}
    \Big[
        \E_{a\sim\pi_x}[q(x,a)]
    \Big],
\]
\[
    \text{where }
    q(x,a)
    :=
    \inf_{Q_{x,a}\in\cU_{\cY}(Q^0_{x,a},\epsilon_{\cY})}
    \E_{y\sim Q_{x,a}}[y]
\]
is the robust reward for context-action cell $(x,a)$.
\end{definition}

The factored objective separates into an inner reward-distribution DRO problem for each context-action cell and an outer context-distribution DRO problem. The next two lemmas give dual representations of these problems. Their proofs follow standard arguments and are included in Appendix~\ref{app:duality} for the reader's convenience.

\begin{lemma}[Reward-level OT duality]
\label{lem:reward_dual}
For every $(x,a)\in\cX\times\cA$,
\[
    q(x,a)
    =
    \sup_{\lambda\in[0,D/\epsilon_{\cY}]}
    \left\{
        -\lambda\epsilon_{\cY}
        +
        \E_{y\sim Q^0_{x,a}}
        \left[
            \inf_{z\in\cY}
            \{z+\lambda c_{\cY}(y,z)\}
        \right]
    \right\}.
\]
\end{lemma}

\begin{lemma}[Context-level OT duality]
\label{lem:context_dual}
For every policy $\pi\in\Pi$,
\[
\begin{aligned}
    V(\pi)
    =
    \sup_{\eta\in[0,D/\epsilon_{\cX}]}
    \left\{
        -\eta\epsilon_{\cX}
        +
        \E_{x\sim P^0}
        \left[
            \inf_{z\in\cX}
            \Big\{
                \E_{a\sim\pi_z}[q(z,a)]
                +
                \eta c_{\cX}(x,z)
            \Big\}
        \right]
    \right\}.
\end{aligned}
\]
\end{lemma}

\subsection{Contributions}
\paragraph{Sharp confidence bounds for robust rewards.}
Subsection~\ref{ssec:reward-lcb} derives a finite-sample confidence bound $\beta_Y$ for the robust reward $q(x,a)$ in each context--action cell. The key observation is that the OT dual integrands $s_\lambda(y)=\inf_{z\in\cY}\{z+\lambda c_{\cY}(y,z)\}$ are pointwise ordered as $\lambda$ varies. This structure lets us control the entire dual family without discretizing either the dual parameter or the reward support. Consequently, the bound applies to arbitrary bounded reward distributions, including continuous ones, and removes the reward-support-cardinality dependence in the bound of \citet{shen2024wasserstein}. Remark~\ref{rem:shen-reward-concentration} shows that it is also strictly sharper.

\paragraph{Uniform confidence bounds for robust policy values.}
Subsection~\ref{ssec:policy-value-lcb} propagates the reward bounds through the context-DRO problem and combines them with a distribution-dependent context-estimation bound, yielding a lower confidence bound $\beta_X$ for $V(\pi)$ simultaneously for every policy $\pi\in\Pi$. Retaining the dependence on $P^0$, rather than replacing it with a bound in terms of $|\cX|$, allows the context-estimation term in the regret analysis to be absorbed into $C_{\rm rob}^\star$.

\paragraph{Pessimistic learning for factored OT-DRO.} 
Subsection \ref{ssec:pessimistic-policy-learning} develops a pessimistic policy learning procedure (Algorithm \ref{alg:opl}) that maximizes the pessimistic value $\underline V(\pi;\delta) = \widehat V(\pi) - \beta_X(\delta)$ over the policy class $\Pi$. The algorithm requires knowledge of neither the true context distribution $P^0$ nor the true behavior policy $\pi^0$. For the unrestricted policy class, a maximizing policy is obtained by choosing $\widehat\pi(x)\in\argmax_{a\in\cA}\underline q(x,a)\forall x$, making the policy optimization computationally simple.

\paragraph{Minimax optimal regret rate.} 
In Section \ref{sec:coverage-rate}, Theorem \ref{thm:coverage-rate} gives a regret bound for the learned policy in terms of the robust optimal-policy coverage coefficient $C_{\rm rob}^\star$ (Assumption \ref{ass:coverage}). Remark \ref{rem:nonrobust-coverage} shows that this coefficient is a natural generalization of the single-policy coverage coefficient of \citet{li2022pessimism} to the factored OT-DRO setting by accounting for the worst-case context shift. Theorem \ref{thm:minimax-lower-bound} in Section \ref{sec:minimax-lower-bound} shows that our regret rate, including its dependence on $C_{\rm rob}^\star$, is minimax-optimal.

\paragraph{Numerical results.} 
Section~\ref{sec:numerical-experiments} checks the predicted confidence and regret behavior in synthetic experiments and evaluates learned policies on logged recommendations and on stroke trial data held out by country. Our learner outperforms existing methods, demonstrating its practical value.

\subsection{Related work}

\paragraph{Distributionally Robust Offline Contextual Bandits.}
Table \ref{tab:regret-comparison} compares the policy-learning guarantees of existing distributionally robust approaches. \citet{si2020distributionally} study a joint KL-DRO objective using self-normalized inverse propensity weighting. They establish asymptotic $n^{-1/2}$ convergence, assuming that the behavior policy is known. \citet{kallus2022doubly} retain the joint KL-DRO objective and develop a doubly robust learner that permits an estimated behavior policy. Their finite-sample regret contains the product $\rho_{\pi^0}\rho_f$ of the behavior-policy and transformed-outcome regression errors. Consequently, the $n^{-1/2}$ rate is retained when both nuisance estimators converge faster than $n^{-1/4}$.

\citet{mu2022factored} introduce separate KL ambiguity sets for context and reward shifts. Their cellwise plug-in analysis requires uniform context and action coverage. The resulting learning rate additionally depends on the reward-layer KL radius $\epsilon_{\cY}$ and a positive lower bound on its optimal dual variable $\lambda^\star$. \citet{shen2024wasserstein} replace the two KL ambiguity sets by factored squared-Wasserstein balls. Their Theorem~5.1 contains a term depending on $|\Xi|$, where $\Xi$ is the finite, known support of a latent observation $\xi$ and the reward is represented as $Y=y_{x,a}(\xi)$. In our direct-reward model, one takes $\xi=Y$ and $y_{x,a}(\xi)=\xi$, so $\Xi=\cY$. In Table~\ref{tab:regret-comparison}, $|\cY|$ refers there to the cardinality of a finite discrete reward support, and their guarantee does not directly cover continuous reward distributions.

Our work follows the factored approaches but accommodates OT ambiguity sets with general transport cost. Moreover, our bound replaces the worst-cell coverage factor $1/\sqrt{\underline P^0\underline\pi^0}$ with the robust optimal-policy coverage coefficient $C_{\rm rob}^\star$. It has no dependence on optimal dual-variable bounds or the reward-support cardinality, and requires no assumed convergence rates for nuisance estimators. The lower bound shows that the dependence on $C_{\rm rob}^\star$ and $\log(|\cX||\cA|/\delta)$ is minimax-optimal.

\begin{table}[ht]
\centering
\renewcommand{\arraystretch}{1.5}
\begin{tabular}{m{0.28\textwidth} m{0.61\textwidth}}
\hline
\textbf{Work} & \textbf{Regret Bound}\\
\hline
\shortstack[l]{\cite{si2020distributionally}\\Joint KL-DRO} &
Asymptotic convergence at $n^{-1/2}$ \\

\shortstack[l]{\cite{kallus2022doubly}\\Joint KL-DRO} &
$\displaystyle
O\!\left(
    \frac{\overline\lambda e^{D/\underline\lambda}}{\underline\pi^0\sqrt n}\left[
        \kappa(\Pi)
        +\frac{D\overline\lambda}{\underline\lambda^2}
        +\sqrt{\log(1/\delta)}\right]
    + \frac{\overline\lambda e^{D/\underline\lambda}}{(\underline\pi^0)^2}\rho_{\pi^0}\rho_f
    \right)$ \\

\shortstack[l]{\cite{mu2022factored}\\Factored KL-DRO} &
$\displaystyle
O\!\left(
    \frac{D^3e^{D/\underline\lambda}}{\epsilon_{\cY}^2\underline\lambda^2\underline P^0\underline\pi^0}
    \sqrt{\frac{\log n\,\log(|\cX||\cA|/\delta)}{n}}
    \right)$ \\

\shortstack[l]{\cite{shen2024wasserstein}\\Factored Wasserstein-DRO} &
$\displaystyle O\!\left(
    D\sqrt{\frac{|\cY|+\log(|\cX||\cA|/\delta)}{n\underline P^0\underline\pi^0}}
    \right)$ \\

\shortstack[l]{\textbf{This work}\\Factored OT-DRO\\(Theorem \ref{thm:coverage-rate})} &
$\displaystyle O\!\left(
    D C_{\rm rob}^\star
    \sqrt{\frac{\log(|\cX||\cA|/\delta)}{n}}\right)$ \\

\shortstack[l]{Lower bound\\(Theorem \ref{thm:minimax-lower-bound})} &
$\displaystyle \Omega\!\left(
    D C_{\rm rob}^\star
    \sqrt{\frac{\log(|\cX||\cA|/\delta)}{n}}\right)$ \\
\hline
\end{tabular}
\caption{Regret bounds for offline contextual bandit learning under distributional shift.\protect\footnotemark}
\label{tab:regret-comparison}
\end{table}
\footnotetext{Written in our notation, with $\underline\pi^0=\min_{x,a}\pi^0_x(a),\quad\underline P^0=\min_x P^0(x),\quad\overline\lambda=\sup\lambda^\star,\quad\underline\lambda=\inf\lambda^\star$. Moreover, $\kappa(\Pi)$ is the Hamming entropy integral, $\rho_{\pi^0}$ is the $L_2$ behavior-policy estimation rate, and $\rho_f$ is the uniform $L_2$ estimation rate of the transformed-outcome regressions.}

\paragraph{Pessimistic Offline Contextual Bandits.}
In the absence of environmental distribution shift, pessimistic offline learning can replace uniform coverage by coverage of an optimal policy. Direct approaches construct lower confidence bounds by estimating conditional rewards. In the tabular setting, \citet{rashidinejad2021bridging} obtain matching upper and lower bounds governed by a single-policy density-ratio coefficient. For linear reward models, \citet{li2022pessimism} derive a sharper instance-dependent regret bound that aggregates inverse behavior probabilities across contexts. These methods rely on tabular or linear outcome structure.

Propensity-weighted approaches instead construct pessimistic estimates of policy-level performance. For general policy classes, \citet{jin2025policy} establish minimax guarantees depending on the overlap of the optimal policy, covering both fixed and adaptively collected logging data, while \citet{wang2024oracle} obtain analogous coverage-adaptive guarantees with an oracle-efficient learner. Recent work by \citet{wan2026risk} extends this approach to Lipschitz risk functionals using propensity-weighted estimators of the policy-induced reward distribution. These methods assume that the logging propensities are known.

All these works evaluate policies under a nominal environment, without adversarial shifts in either the conditional reward laws or the context distribution. Our method extends the pessimistic direct approach to the factored OT-DRO objective, which accounts separately for shifts in the context distribution and the conditional reward distributions. Correspondingly, $C_{\rm rob}^\star$ extends the single-policy coverage principle by incorporating the worst-case context distribution in the ambiguity set. In the no-context-shift specialization $\epsilon_{\cX}=0$, it reduces to the corresponding non-robust coverage scaling; see Remark~\ref{rem:nonrobust-coverage}.

\section{Pessimistic Policy Learning for Factored OT-DRO}

The learning procedure follows three steps. First, it estimates the robust reward $\widehat q(x,a)$ in every observed cell, subtracts its confidence penalty $\beta_Y(N_{x,a};\delta)$, and clips at zero to obtain the pessimistic reward $\underline q(x,a)$. Second, it evaluates policies through the outer context-DRO problem using these pessimistic rewards. Third, it selects the policy with the largest pessimistic empirical value. A context-distribution penalty $\beta_X(\delta)$ turns this objective into a valid policy-value lower confidence bound. Because this penalty is common to every policy, it provides the lower-confidence interpretation without affecting policy selection. The next three subsections develop these steps in order.

\paragraph{Empirical distributions.}
All estimators use the full logged dataset $\cD_n$. For each cell $(x,a)$, define the sample count $N_{x,a}:=\sum_{i=1}^n\1\{X_i=x,A_i=a\}$. The empirical context distribution is denoted by $\widehat P(x):=\frac1n\sum_{i=1}^n\1\{X_i=x\}$.

\subsection{Robust Reward Lower Confidence Bounds}
\label{ssec:reward-lcb}
For $N_{x,a}>0$, define the empirical robust reward by
\begin{equation}
\label{eq:hatq}
    \widehat q(x,a)
    :=
    \sup_{\lambda\in[0,D/\epsilon_{\cY}]}
    \left\{
        -\lambda\epsilon_{\cY}
        +
        \frac1{N_{x,a}}
        \sum_{i:X_i=x,A_i=a}
        \inf_{z\in\cY}\{z+\lambda c_{\cY}(Y_i,z)\}
    \right\}.
\end{equation}
If $N_{x,a}=0$, define $\widehat q(x,a):=0$.

For $m\in\{0,1,\dots,n\}$, define the cellwise reward confidence radius
\begin{equation}
\label{eq:beta_Y}
    \beta_Y(m;\delta)
    :=
    \begin{cases}
    D, & m=0,\\[0.4em]
    D\min\left\{
        1,
        \sqrt{\dfrac{\pi}{2m}}
        +
        \sqrt{
            \dfrac{\log(2|\cX||\cA|/\delta)}{2m}
        }
    \right\},
    & m\ge1.
    \end{cases}
\end{equation}
Now define the pessimistic reward estimate
\[
    \underline q(x,a)
    :=
    \max\{0,\ \widehat q(x,a)-\beta_Y(N_{x,a};\delta)\}.
\]
Thus $\underline q(x,a)\in[0,D]$ for every cell; in particular, an empty cell has $\widehat q(x,a)=\underline q(x,a)=0$. The next proposition shows that $\underline q$ is a valid lower confidence bound for the robust reward $q$.
\begin{proposition}[Reward-cell concentration]
\label{prop:reward_concentration}
Under Assumptions \ref{ass:logged-model} and \ref{ass:transport-cost}, with probability at least $1-\delta/2$ over the draw of the logged data $\cD_n$, simultaneously for every $(x,a)\in\cX\times\cA$,
\[
    |\widehat q(x,a)-q(x,a)|
    \le
    \beta_Y(N_{x,a};\delta).
\]
On the same event,
\(
    0\le\underline q(x,a)\le q(x,a)
    \;
    \forall (x,a),
\)
and
\(
    q(x,a)-\underline q(x,a)
    \le
    2\beta_Y(N_{x,a};\delta)
    \;
    \forall (x,a).
\)
\end{proposition}

\begin{remark}[Comparison with \cite{shen2024wasserstein}]
\label{rem:shen-reward-concentration}
For each context--action cell, \citet{shen2024wasserstein} use the plug-in empirical conditional law and control the difference between the empirical and population Wasserstein-robust costs by the total-variation error of that law on a finite, discrete reward support $\cY$; see their Lemma~A.4. Applied to a cell containing $m$ observations, that route gives a confidence radius of
\[
    D\left\{
        \sqrt{\frac{2|\cY|}{m}}
        +
        \sqrt{\frac{4\log(2|\cX||\cA|/\delta)}{m}}
    \right\},
\]
which becomes a dependence on the reward-support cardinality. Our proof of Proposition~\ref{prop:reward_concentration} instead controls the OT dual integrand $s_\lambda(y)=\inf_{z\in\cY}\{z+\lambda c_{\cY}(y,z)\}$ directly. Since $s_\lambda(y)$ increases in $\lambda$ pointwise in $y$, its pointwise ordering, a layer-cake representation, and the Dvoretzky--Kiefer--Wolfowitz inequality yield the radius
\[
    D\left\{
        \sqrt{\frac{\pi}{2m}}
        +
        \sqrt{\frac{\log(2|\cX||\cA|/\delta)}{2m}}
    \right\}
\]
before truncation at $D$; see our detailed proof in Appendix~\ref{app:proof-reward-concentration}. Thus our bound has no reward-support-cardinality term. It therefore applies to arbitrary bounded reward laws, including continuous ones, and is strictly sharper. This improvement comes from the ordered-class analysis.
\end{remark}

\subsection{From Reward Bounds to Policy-Value Bounds}
\label{ssec:policy-value-lcb}
With the empirical context distribution $\widehat P$ and the pessimistic reward table $\underline q$, the empirical context-robust value based on pessimistic rewards is defined by
\begin{equation}
\label{eq:hatV}
\begin{aligned}
    \widehat V(\pi)
    &:={}
    \inf_{P\in\cU_{\cX}(\widehat P,\epsilon_{\cX})}
    \E_{X\sim P}\left[
        \sum_{a\in\cA}\pi_X(a)\underline q(X,a)
    \right] \\
    &={}
    \sup_{\eta\in[0,D/\epsilon_{\cX}]}
    \left\{
        -\eta\epsilon_{\cX}
        +
        \E_{X\sim \widehat P}
        \left[
            \min_{z\in\cX}
            \left\{
                \sum_{a\in\cA}\pi_z(a)\underline q(z,a)
                +
                \eta c_{\cX}(X,z)
            \right\}
        \right]
    \right\}.
\end{aligned}
\end{equation}
Define the distribution-dependent context penalty
\begin{equation}
\label{eq:beta_X}
    \beta_X(\delta)
    :=
    D
    \min\left\{
        1,
        \frac{1}{2\sqrt n}
        \left[
            \sum_{x\in\cX}\sqrt{P^0(x)(1-P^0(x))}
            +\sqrt{2\log(2/\delta)}
        \right]
        \right\}.
\end{equation}
The pessimistic factored OT-DRO value estimate of policy $\pi$ is
\[
    \underline V(\pi;\delta)
    :=
    \widehat V(\pi)-\beta_X(\delta).
\]
The following proposition shows that $\underline V(\pi;\delta)$ is a valid lower confidence bound for the factored OT-DRO value $V(\pi)$. Its proof is given in Appendix~\ref{app:proof-value-lcb}.
\begin{proposition}[Policy-value lower confidence bound]
\label{prop:value_lcb}
Under Assumption \ref{ass:logged-model} and \ref{ass:transport-cost}, with probability at least $1-\delta$ over the draw of the logged data $\cD_n$, simultaneously for every $\pi\in\Pi$,
\(
    \underline V(\pi;\delta)
    \le
    V(\pi).
\)
Consequently, on the same event,
\[
    V(\pi)-\underline V(\pi;\delta)
    \le
    2\sup_{P\in\cU_{\cX}(P^0,\epsilon_{\cX})}
    \E_{X\sim P}
    \left[
        \sum_{a\in\cA}
        \pi_X(a)
        \beta_Y(N_{X,a};\delta)
    \right]
    +
    2\beta_X(\delta)
    \qquad
    \forall \pi\in\Pi.
\]
\end{proposition}

\subsection{Pessimistic Policy Learning}
\label{ssec:pessimistic-policy-learning}
Pessimistic Policy Learning maximizes the pessimistic value $\underline V(\pi;\delta)$ over the policy class $\Pi$. Because the context penalty $\beta_X(\delta)$ is independent of $\pi$, this is equivalent to maximizing the empirical value $\widehat V(\pi)$:
\(
    \argmax_{\pi\in\Pi}\underline V(\pi;\delta)
    =
    \argmax_{\pi\in\Pi}\widehat V(\pi).
\)
Consequently, $\beta_X(\delta)$ need not be computed. There is no need to replace the sharp $P^0$-dependent penalty by a looser computable one: the learner maximizes $\widehat V$, while $\underline V$ remains a valid policy-value lower confidence bound in the analysis. Algorithm~\ref{alg:opl} summarizes the resulting procedure.

\begin{algorithm}[ht]
\caption{Pessimistic Factored OT-DRO Offline Policy Learning}
\label{alg:opl}
\begin{algorithmic}[1]
\STATE Input: data $\cD_n$, policy class $\Pi$, costs $c_{\cY},c_{\cX}$, radii $\epsilon_{\cY},\epsilon_{\cX}$, reward bound $D$, confidence $\delta$.
\STATE Compute $\widehat P$ and the cell counts $N_{x,a}$.
\FOR{each $(x,a)\in\cX\times\cA$}
    \STATE Compute $\widehat q(x,a)$ by~\eqref{eq:hatq} and $\beta_Y(N_{x,a};\delta)$ by~\eqref{eq:beta_Y}.
    \STATE Compute $\underline q(x,a)=\max\{0,\widehat q(x,a)-\beta_Y(N_{x,a};\delta)\}$.
\ENDFOR
\STATE Solve $\max_{\pi\in\Pi}\widehat V(\pi)$, with $\widehat V(\pi)$ defined by~\eqref{eq:hatV}.
\STATE Output any $\widehat\pi_\delta\in\argmax_{\pi\in\Pi}\widehat V(\pi)$; equivalently, $\widehat\pi_\delta\in\argmax_{\pi\in\Pi}\underline V(\pi;\delta)$.
\end{algorithmic}
\end{algorithm}

\paragraph{Lexicographic tie-breaking with unclipped pessimistic rewards.}
In Algorithm~\ref{alg:opl}, when the primary optimization has multiple solutions, we may break ties using the corresponding unclipped scores $\widetilde q(x,a):=\widehat q(x,a)-\beta_Y(N_{x,a};\delta)$. Specifically, define
\[
    \widetilde V(\pi)
    :=
    \inf_{P\in\cU_{\cX}(\widehat P,\epsilon_{\cX})}
    \E_{X\sim P}
    \left[
        \sum_{a\in\cA}\pi_X(a)\widetilde q(X,a)
    \right],
    \qquad
    \widehat\Pi_\delta
    :=
    \argmax_{\pi\in\Pi}\widehat V(\pi),
\]
and select $\widehat\pi_\delta \in \argmax_{\pi\in\widehat\Pi_\delta}\widetilde V(\pi)$.
This secondary criterion retains information discarded by clipping: when many pessimistic rewards $\underline q$ equal zero, it favors, among the policies tied under the clipped objective, those with larger values of $\widehat q(x,a)-\beta_Y(N_{x,a};\delta)$. Since the selected policy remains an element of $\widehat\Pi_\delta$, this tie-breaking rule does not alter any of the theoretical guarantees.

\paragraph{Optimization over the unrestricted policy class.}
For the class of all stochastic policies, $\Pi=\Delta(\cA)^{|\cX|}$, the optimization in Algorithm~\ref{alg:opl} separates across contexts. An
optimal policy can therefore be chosen deterministically according to $\widehat\pi_\delta(x) \in \argmax_{a\in\cA}\underline q(x,a), \forall x\in\cX$. Indeed, this choice maximizes the pessimistic reward at every context and hence its expectation under every context distribution in the ambiguity set. Moreover, one may choose $\widehat\pi_\delta(x) \in \argmax_{a\in\cA}\widetilde q(x,a), \forall x\in\cX$. Because $\underline q(x,a)=\max\{0,\widetilde q(x,a)\}$, every maximizer of $\widetilde q(x,\cdot)$ is also a maximizer of $\underline q(x,\cdot)$. This implements the lexicographic tie-breaking described above.

\paragraph{Optimization over a structured policy class.}
When $\Pi$ is described by linear constraints on the policy probabilities, the policy-learning step is a linear program; other structured classes may be handled using an appropriate optimization oracle. Exact optimization is not required for the statistical guarantee: if the returned policy satisfies
\(
\widehat V(\widehat\pi_\delta)
\ge
\sup_{\pi\in\Pi}\widehat V(\pi)-\tau
\)
for some $\tau\ge0$, then all subsequent regret upper bounds hold with an additional additive term $\tau$.

\paragraph{Regret in terms of random cell counts.}
Let the true optimal policy be $\pi^\star\in\argmax_{\pi\in\Pi}V(\pi)$ and the selected policy be $\widehat\pi_\delta\in\argmax_{\pi\in\Pi}\underline V(\pi;\delta)$. On the event of Proposition~\ref{prop:value_lcb}, we have $V(\widehat\pi_\delta)\ge\underline V(\widehat\pi_\delta;\delta)\ge\underline V(\pi^\star;\delta)$. Applying the gap bound from Proposition~\ref{prop:value_lcb} gives a regret bound in terms of the random cell counts:
\[
    V(\pi^\star)-V(\widehat\pi_\delta)
    \le
    2\sup_{P\in\cU_{\cX}(P^0,\epsilon_{\cX})}
    \E_{X\sim P}
    \left[
        \sum_{a\in\cA}
        \pi^\star_X(a)
        \beta_Y(N_{X,a};\delta)
    \right]
    +2\beta_X(\delta).
\]
The next section converts this into a deterministic regret rate under a coverage condition.

\section{Regret Under Robust Optimal-Policy Coverage}
\label{sec:coverage-rate}
The preceding bound depends on random cell counts. We now convert it into a deterministic rate under a robust optimal-policy coverage condition. Assumption~\ref{ass:coverage} generalizes the single-policy coverage condition to the factored OT-DRO setting. Theorem~\ref{thm:coverage-rate} then gives an explicit regret bound in terms of this robust optimal-policy coverage coefficient.
\begin{assumption}[Robust optimal-policy coverage]
\label{ass:coverage}
\[
    C_{\rm rob}^\star
    :=
    \sup_{P\in \cU_{\cX}(P^0,\epsilon_{\cX})}
    \sum_{x:P(x)>0}
    \frac{P(x)}{\sqrt{P^0(x)}}
    \sum_{a:\pi^\star_x(a)>0}
    \frac{\pi^\star_x(a)}{\sqrt{\pi^0_x(a)}}
    <\infty.
\]
\end{assumption}

\begin{remark}[Recovery of non-robust single-policy coverage]
\label{rem:nonrobust-coverage}
Consider the no-context-shift specialization $\epsilon_{\cX}=0$ with
$\cU_{\cX}(P^0,0)=\{P^0\}$; the singleton condition holds, for example,
when $c_{\cX}(x,z)>0$ for all $x\ne z$. Then
\[
\begin{aligned}
    C_{\rm rob}^\star\big|_{\epsilon_{\cX}=0}
    &=
    \sum_{x:P^0(x)>0}\sqrt{P^0(x)}
    \sum_{a:\pi^\star_x(a)>0}
        \frac{\pi_x^\star(a)}{\sqrt{\pi_x^0(a)}}\\
    &=
    \sum_{x:P^0(x)>0}
    \sqrt{\frac{P^0(x)}{\pi_x^0(\pi^\star(x))}}
    \,\text{ for a deterministic optimal policy}.
\end{aligned}
\]
This is exactly the instance-dependent coverage term in Corollary 1 of \citet{li2022pessimism}. Meanwhile, the density-ratio single-policy concentrability coefficient of \citet{rashidinejad2021bridging} gives the coarser comparison
\[
    C_{\rm rob}^\star\big|_{\epsilon_{\cX}=0}
    \le
    \sqrt{|\cX|C_1^\star},
    \qquad
    C_1^\star
    :=
    \max_{x:P^0(x)>0}
    \frac{1}{\pi_x^0(\pi^\star(x))},
\]
resulting in regret bounds of order $\tilde O(\sqrt{|\cX|C^\star_1/n})$ in their Theorem 4 (upper bound) and Theorem 5 (lower bound). Similarly, the stochastic-policy pseudo-loss of \citet{wang2024oracle} also gives
\[
    C_{\rm rob}^\star\big|_{\epsilon_{\cX}=0}
    \le
    \sqrt{|\cX|C_2^\star},
    \qquad
    C_2^\star
    :=
    \E_{X\sim P^0}
    \left[
        \sum_{a:\pi_X^\star(a)>0}
        \frac{\pi_X^\star(a)}{\pi_X^0(a)}
    \right].
\]
Hence the no-context-shift specialization recovers the familiar single-policy coverage scaling. Note that the factor $\sqrt{|\cX|}$ in this comparison comes from replacing the instance-dependent sum by the coarser worst-case coefficient $C_1^\star$; it is not an additional cost of our robust bound.
\end{remark}

\paragraph{Robust optimal-policy coverage is stronger}
Because $P^0\in\cU_{\cX}(P^0,\epsilon_{\cX})$, the robust coverage coefficient is at least as large as the corresponding coefficient evaluated under the nominal context distribution. Moreover, since $\epsilon_{\cX}>0$ and $c_{\cX}$ is bounded, every context $x\in\cX$ receives positive mass under some $P\in\cU_{\cX}(P^0,\epsilon_{\cX})$: one can transport an arbitrarily small positive amount of mass from any context in the support of $P^0$ to $x$ while remaining within the transport budget. Consequently, finiteness of $C_{\rm rob}^\star$ is equivalent to
\[
    P^0(x)>0
    \quad\text{for every }x\in\cX,
    \qquad\text{and}\qquad
    \pi_x^0(a)>0
    \quad\text{for every }(x,a)\text{ such that }\pi_x^\star(a)>0.
\]
Thus, unlike the no-context-shift setting, which requires coverage only at contexts in the support of $P^0$, positive-radius context robustness requires the logged data distribution to have full context support. Nevertheless, it requires action coverage only along the optimal robust policy and imposes no coverage requirement on actions that $\pi^\star$ never uses. The mismatch between the contexts emphasized by an adversarial deploying distribution and those observed under the logging distribution is what makes robust optimal-policy coverage stronger than its non-robust counterpart.

Using the robust optimal-policy coverage coefficient, we can convert the random-cell-count regret bound into a deterministic rate in the following theorem. Its proof is given in Appendix~\ref{app:learning}.
\begin{theorem}[Regret under robust optimal-policy coverage]
\label{thm:coverage-rate}
Let $\delta\in(0,1)$ and suppose Assumption~\ref{ass:logged-model}-\ref{ass:coverage} holds. Then the learner run with confidence $\delta/2$ satisfies, with probability at least $1-\delta$,
\[
V(\pi^\star)-V(\widehat\pi_{\delta/2})
\le
\left(
2\sqrt{\frac{2}{1-\log 2}}
+1+\sqrt2
\right)
D C_{\rm rob}^\star
\sqrt{\frac{\log(4|\cX||\cA|/\delta)}{n}}.
\]
\end{theorem}
In our analysis, the context-estimation error, controlled by the distribution-dependent radius $\beta_X(\delta)$ in \eqref{eq:beta_X}, is absorbed into the robust optimal-policy coverage coefficient:
\[
\sum_{x\in\cX}\sqrt{P^0(x)(1-P^0(x))}
\le
\sum_{x\in\cX}\sqrt{P^0(x)}
\le
C_{\rm rob}^\star.
\]
By comparison, Theorem~5.1 of \citet{shen2024wasserstein}, using the total-variation bound in their equation~(A.3), incurs a context-estimation term of order $D(\sqrt{\log(1/\delta)/n}+\sqrt{|\cX|/n})$. Under their uniform context-coverage assumption, $|\cX|\le 1/\underline P^0$, so the second term can be absorbed into the overall $1/\sqrt{\underline P^0}$ dependence.

\section{Minimax Lower Bound}
\label{sec:minimax-lower-bound}

We next show that the leading dependence on robust optimal-policy coverage and the full context--action cardinality logarithm are unavoidable. The construction has one anchor and $|\cX|-1$ hard contexts, and it indexes the hard family jointly by a context and an action. Both ambiguity radii are strictly positive: the context adversary transports mass from the anchor toward the hard contexts, while the reward adversary lowers every Bernoulli reward law within a squared-Wasserstein ball. The proof is given in Appendix~\ref{app:lower-bound}.

\begin{theorem}[Lower bound]
\label{thm:minimax-lower-bound}
Fix $D,\kappa>0$, finite sets $\cX$ and $\cA$ with $|\cX|\ge3$ and $|\cA|\ge3$, $0<\epsilon_{\cX}<\kappa/2$, and $0<\epsilon_{\cY}\le D^2/2$. Write
\[
    r:=|\cX|-1,
    \qquad
    L:=|\cA|-1,
    \qquad
    \rho:=\frac{\epsilon_{\cX}}{\kappa},
    \qquad
    w:=\frac{1}{2r}+\rho.
\]
Fix $0<\mu\le\frac{w^2}{1+Lw^2}$. For every $\delta\in(0,1/4]$ satisfying $n\mu\ge\frac{7r}{8}\log\left(\frac{4|\cX||\cA|}{\delta}\right)$, and every learner $\widehat\pi$ that outputs a policy in the supplied policy class, there exists a $|\cX|$-context, $|\cA|$-action instance with both context and reward shifts active and
\[
    C_{\rm rob}^\star
    =
    w\sqrt{\frac{2r}{\mu}}
    +
    \left(\frac12-\frac{1}{2r}\right)
        \sqrt{\frac{2r}{1-L\mu}}
    +
    \left(\frac12-\rho\right)\sqrt2,
\]
such that, under its logged-data law $\Pbb^n$,
\[
    \Pbb^n\left(
        V(\pi^\star)-V(\widehat\pi)
        \ge
        \frac{\sqrt7}{512}\,
        D C_{\rm rob}^\star
        \sqrt{\frac{\log(4|\cX||\cA|/\delta)}{n}}
    \right)
    >\delta.
\]
\end{theorem}

Theorem~\ref{thm:minimax-lower-bound} matches the rate of Theorem~\ref{thm:coverage-rate}, up to universal constants. Thus the reward scale, robust optimal-policy coverage, context--action cardinality logarithm, confidence dependence, and $n^{-1/2}$ rate are all unavoidable in the worst case. 

\section{Numerical Experiments}
\label{sec:numerical-experiments}

\subsection{Synthetic experiments}

\paragraph{Robust-reward confidence bound.}
Appendix~\ref{app:experiment-q-confidence} evaluates the bound in Proposition~\ref{prop:reward_concentration} on $58$ random and structured reward distributions, using quadratic transport cost and $20$ sample sizes between $10$ and $10^6$. No bound violation occurs across $1.16$ million replications. The empirical $95$th-percentile estimation errors exhibit approximately $m^{-1/2}$ decay and remain below the theoretical radius throughout (Figure~\ref{fig:q-confidence-bound}). The radius is at least $2.79$ times these empirical quantiles, supporting the confidence guarantee on the tested distributions while demonstrating conservative constants.

\paragraph{Dependence on sample size and coverage.}
Appendix~\ref{app:experiment-regret-n-c} studies Bernoulli instances with 4 contexts, 9 actions, and an exactly prescribed coverage coefficient $C_{\rm rob}^\star$. At each of $425$ combinations of $n$ and $C_{\rm rob}^\star$, we select a reward gap with large estimated regret and evaluate it using independent simulations. The regret curves exhibit a small-sample plateau followed by decay (Figures~\ref{fig:regret-vs-n}--\ref{fig:regret-normalized}). For $n\ge 2^{22}$, a joint log--log fit gives exponents $-0.5004$ for $n$ and $1.0021$ for $C_{\rm rob}^\star$. These results support the $C_{\rm rob}^\star/\sqrt n$ dependence predicted by Theorem~\ref{thm:coverage-rate} in the large-sample regime of this family.

\paragraph{Rare suboptimal actions.}
Appendix~\ref{app:experiment-rare-suboptimal} reduces the logging probability of a suboptimal action while fixing $C^\star_{\rm rob}=64$. As the worst-cell coverage factor rises from $64$ to $4{,}096$, mean robust regret falls by $3.8\%$--$7.6\%$ across the three tested sample sizes.

\subsection{Open Bandit Dataset}
\paragraph{Dataset.}
We use the Open Bandit Dataset (OBD), collected on the fashion e-commerce platform ZOZOTOWN \citep{saito2020open}. The uniform-random logs for the \emph{men}, \emph{women}, and \emph{all} campaigns contain approximately $0.45$--$1.37$ million rows per campaign; recommended items are actions, and binary clicks are rewards. We treat each logged item as a separate bandit observation and split each log by rows into $60\%$ training, $20\%$ validation, and $20\%$ test data. Appendix~\ref{app:experiment-obd} provides the full protocol.

\paragraph{Protocol.}
We implement Algorithm~\ref{alg:opl} over the unrestricted policy class. To adapt the ambiguity radius to the low click-through rate (CTR), we set
\(
    \epsilon_{\cY}
    =\rho\,\widehat{\mathrm{CTR}}_{\rm train}
\)
where $\rho$ is a relative-radius multiplier. Motivated by the conservatism observed above, we introduce a confidence scale $c\ge0$ and select
\(
    \widehat\pi_c(x)\in
    \argmax_{a\in\cA}
    \left\{
        \widehat q(x,a)
        -c\,\beta_Y(N_{x,a};0.1)
    \right\}.
\)
Using the unclipped scores implements the tie-breaking described in Section~\ref{ssec:pessimistic-policy-learning}. For this policy class, selection is independent of $\epsilon_\cX$, and $\beta_X$ need not be computed. The choice $c=1$ retains the theoretical confidence radius; choosing $c<1$ reduces conservatism. We tune $(\rho,c)$ over a $36$-point grid using validation self-normalized inverse propensity scoring. The selected policy is then evaluated on the held-out test split.

\paragraph{Results.}
Table~\ref{tab:obd-main} compares our learner with factored W-DRO \citep{shen2024wasserstein}, the IPW Learner \citep{saito2020open}, POEM \citep{swaminathan2015counterfactual}, and joint KL-DRO \citep{si2020distributionally}. Our learner attains the largest test SNIPS point estimate in every campaign.

\begin{table}[ht]
    \centering
    \small
    \setlength{\tabcolsep}{5pt}
    \begin{tabular}{lrrrrrr}
    \hline
    Campaign & \shortstack{Behavior\\policy} & \shortstack{Factored\\W-DRO} & \shortstack{IPW\\Learner} & POEM & \shortstack{Joint\\KL-DRO} & Ours \\
    \hline
    men   & $0.544\%$ & $2.129\%$ & $1.087\%$ & $1.570\%$ & $0.614\%$ & $\mathbf{2.155\%}$ \\
    women & $0.483\%$ & $1.763\%$ & $1.410\%$ & $1.315\%$ & $0.936\%$ & $\mathbf{1.796\%}$ \\
    all   & $0.327\%$ & $1.734\%$ & $1.436\%$ & $1.086\%$ & $0.837\%$ & $\mathbf{1.817\%}$ \\
    \hline
    \end{tabular}
    \caption{Test policy-value estimates (\%) on OBD. Learned policies are evaluated by SNIPS.}
    \label{tab:obd-main}
\end{table}

\subsection{International Stroke Trial}
\label{ssec:ist-main}

\paragraph{Dataset and geographic split.}
The International Stroke Trial (IST) randomized $19{,}435$ patients with suspected acute ischemic stroke to six combinations of aspirin (given or avoided) and heparin (none, low dose, or higher dose) \citep{ist1997trial,sandercock2011ist}. We define four contexts by age ($<70$ or $\geq70$) and baseline consciousness (fully conscious or impaired). The four contexts are unevenly represented, with shares of $6.5\%$ to $46.1\%$ (Table~\ref{tab:ist-context-action-counts}). The reward is an indicator of being alive and independent at six months. We reserve Australia, Austria, Czechia, and the Netherlands for testing ($2{,}002$ patients with observed outcomes) and use the other countries for training and validation. This geographic split creates a plausible distributional shift. The empirical context distributions of the four test countries have total-variation distances of $0.031$ to $0.187$ from the pooled training distribution.

\paragraph{Tuning and evaluation.}
We use a common policy class that caps the average probability of assigning either dose of heparin at $30\%$ under equal weights over the four contexts. This constraint couples decisions across contexts, so the context cost and radius can affect the optimizer. We compare $36$ combinations of context cost, context radius multiplier, and reward radius. Each context radius scales a reference calibrated from training-country context distributions. We hold out three groups of non-test countries in turn and average augmented inverse propensity weighting (AIPW) estimates equally over the $13$ validation countries. The selected setting uses a normalized Hamming cost on age and consciousness, a context multiplier of $2$ ($\epsilon_{\cX}=0.293$ after refitting), and $\epsilon_{\cY}=0.05$ under absolute reward cost. Its validation AIPW estimate is $46.62\%$. We tune the IPW Learner, POEM, and joint KL-DRO on the same folds and refit all policies on the non-test countries before scoring the test data. The zero-radii control uses the same pessimistic learner with both ambiguity radii set to zero. Appendix~\ref{app:experiment-ist} gives the full protocol and validation results.

\paragraph{Test results.}
Table~\ref{tab:ist-main} reports equal-country AIPW estimates using the trial's nominal allocation shares as working propensities. OT-DRO has the highest estimate ($43.36\%$), $7.06$ percentage points above the zero-radii control. All four paired $90\%$ intervals lie above zero.

\begin{table}[ht]
    \centering
    \small
    \begin{tabular}{lrr}
    \hline
    Policy & Test AIPW (\%) & OT-DRO improvement \\
    \hline
    OT-DRO             & $\mathbf{43.36}$ $[38.22,50.30]$ & --- \\
    Zero-radii control & $36.30$ $[31.58,43.50]$ & $7.06$ $[1.24,11.45]$ \\
    IPW Learner        & $35.07$ $[30.26,40.81]$ & $8.29$ $[2.51,14.91]$ \\
    POEM               & $35.07$ $[30.26,40.81]$ & $8.29$ $[2.51,14.91]$ \\
    Joint KL-DRO       & $35.84$ $[31.90,42.21]$ & $7.52$ $[2.30,11.05]$ \\
    \hline
    \end{tabular}
    \caption{Held-out IST policy values and paired differences. Brackets give $90\%$ intervals from resampling hospitals within each fixed test country. IPW Learner and POEM yield the same refitted policy up to numerical precision.}
    \label{tab:ist-main}
\end{table}

\section{Conclusion}
\label{sec:conclusion}

We established minimax-optimal regret guarantees for pessimistic policy learning under factored OT-DRO. The guarantees require action coverage only along an optimal robust policy, and the robust optimal-policy coverage coefficient captures how admissible context shifts amplify limited coverage in the logged data. Our analysis combines sharp confidence bounds for robust rewards with a distribution-dependent context-estimation bound. The Open Bandit Dataset illustrates the unrestricted learner on logged recommendations, and IST illustrates the constrained learner when the patient mix changes across countries. These applications estimate policy values on held-out data. Extending the theory beyond finite contexts is a direction for future work.

\subsection*{AI use statement}
In this work, we used generative AI tools to assist with implementing the experimental evaluation, including coding the experiments reported in the paper. We also used generative AI tools to polish writing and to identify related work. We did not use generative AI tools to develop the theoretical model or conceptual framework, formulate mathematical claims, provide ingredients for or write mathematical proofs, propose or refine hypotheses, design the research methodology or experiments, generate or analyze datasets, or interpret experimental results.

All AI-generated experimental code was reviewed, verified, and tested for correctness by the authors. AI-identified related work was independently reviewed and validated by the authors. We reviewed all AI-assisted work and take responsibility for the final content of this work, including text, claims, code, and other artifacts produced with the aid of generative AI.

\bibliography{references}
\bibliographystyle{plainnat}

\newpage
\appendix

\section*{Appendix}
\phantomsection
\label{app:appendix}
This appendix provides detailed proofs of the main results and further information on the numerical experiments. The appendix is organized as follows.

\subsection*{Contents}
\phantomsection
\begin{center}
\small
\renewcommand{\arraystretch}{1.7}
\begin{tabular}{@{}p{0.85\linewidth}r@{}}
\hline
\textbf{Item} & \textbf{Page} \\
\hline
\textbf{Appendix~\ref{app:duality}: OT Duality and Deterministic Properties} & \pageref{app:duality} \\
\hspace*{1em}\ref{app:proof-reward-dual}\quad Reward-level OT duality (Lemma~\ref{lem:reward_dual}) & \pageref{app:proof-reward-dual} \\
\hspace*{1em}\ref{app:proof-reward-score-regular}\quad Reward dual integrand ordering and regularity (Lemma~\ref{lem:reward_score_regular}) & \pageref{app:proof-reward-score-regular} \\
\hspace*{1em}\ref{app:proof-context-dual}\quad Context-level OT duality (Lemma~\ref{lem:context_dual}) & \pageref{app:proof-context-dual} \\
\hspace*{1em}\ref{app:proof-context-monotone}\quad Monotonicity and Lipschitzness of the context robust value (Lemma~\ref{lem:context_monotone}) & \pageref{app:proof-context-monotone} \\
\textbf{Appendix~\ref{app:confidence}: Finite-Sample Confidence Bounds} & \pageref{app:confidence} \\
\hspace*{1em}\ref{app:proof-reward-concentration}\quad Reward-cell concentration (Proposition~\ref{prop:reward_concentration}) & \pageref{app:proof-reward-concentration} \\
\hspace*{1em}\ref{app:proof-context-concentration}\quad Total-variation context concentration (Lemma~\ref{lem:context_concentration}) & \pageref{app:proof-context-concentration} \\
\hspace*{1em}\ref{app:proof-value-lcb}\quad Policy-value lower confidence bound (Proposition~\ref{prop:value_lcb}) & \pageref{app:proof-value-lcb} \\
\textbf{Appendix~\ref{app:learning}: Policy-Learning Guarantees} & \pageref{app:learning} \\
\hspace*{1em}\ref{app:proof-count-to-confidence}\quad Count-to-confidence conversion (Lemma~\ref{lem:count-to-confidence}) & \pageref{app:proof-count-to-confidence} \\
\hspace*{1em}\ref{app:proof-coverage-rate}\quad Regret under robust optimal-policy coverage (Theorem~\ref{thm:coverage-rate}) & \pageref{app:proof-coverage-rate} \\
\textbf{Appendix~\ref{app:lower-bound}: Proof of the Minimax Lower Bound (Theorem~\ref{thm:minimax-lower-bound})}& \pageref{app:lower-bound} \\
\hspace*{1em}\ref{app:lower-bound-hard-instances}\quad Step 1: Hard instances & \pageref{app:lower-bound-hard-instances} \\
\hspace*{1em}\ref{app:lower-bound-regret-coverage}\quad Step 2: Robust regret and coverage & \pageref{app:lower-bound-regret-coverage} \\
\hspace*{1em}\ref{app:lower-bound-relative-entropy}\quad Step 3: Relative entropy & \pageref{app:lower-bound-relative-entropy} \\
\hspace*{1em}\ref{app:lower-bound-testing}\quad Step 4: Fano and binary testing & \pageref{app:lower-bound-testing} \\
\textbf{Appendix~\ref{app:experiments}: Experiment Details} & \pageref{app:experiments} \\
\hspace*{1em}\ref{app:experiment-q-confidence}\quad Empirical Behavior of the Robust-Reward Confidence Radius & \pageref{app:experiment-q-confidence} \\
\hspace*{1em}\ref{app:experiment-regret-n-c}\quad Dependence of Regret on $n$ and $C^*_{\rm rob}$ & \pageref{app:experiment-regret-n-c} \\
\hspace*{1em}\ref{app:experiment-rare-suboptimal}\quad Sensitivity to a Rare Suboptimal Action & \pageref{app:experiment-rare-suboptimal} \\
\hspace*{1em}\ref{app:experiment-obd}\quad Real-Data Application: Off-Policy Learning on the Open Bandit Dataset & \pageref{app:experiment-obd} \\
\hspace*{1em}\ref{app:experiment-ist}\quad Geographic Holdout on the International Stroke Trial & \pageref{app:experiment-ist} \\
\hline
\end{tabular}
\end{center}
\bigskip

\input{appendix_duality.tex}
\newpage
\input{appendix_lcb.tex}
\newpage
\input{appendix_learning.tex}
\newpage
\input{appendix_lowerbound.tex}
\newpage
\input{appendix_experiments.tex}

\end{document}

%% file: appendix_duality.tex
\section{OT Duality and Deterministic Properties}
\label{app:duality}

We derive the reward- and context-level dual representations by expressing each transport constraint through couplings and introducing a Lagrange multiplier. We then establish two properties used in the statistical analysis: ordering and regularity of the reward dual integrands, and monotonicity and Lipschitzness of the context-robust value. The proofs use standard arguments and are included for the reader's convenience.

\subsection{Reward-level OT duality (Lemma~\ref{lem:reward_dual})}
\label{app:proof-reward-dual}

\begin{proof}
Fix a context-action pair $(x,a)$ and write $c:=c_{\cY}$. Recall that $q(x,a) = \inf_{Q:\cT_{\cY}(Q^0_{x,a},Q)\le\epsilon_{\cY}} \E_{Z\sim Q}[Z].$
We first write the problem using couplings. Let
\[
    \mathfrak C(Q^0_{x,a})
    :=
    \{\gamma\in\Delta(\cY\times\cY):\gamma_1=Q^0_{x,a}\},
\]
where $\gamma_1$ is the first marginal. If $\gamma\in\mathfrak C(Q^0_{x,a})$, then its second marginal $\gamma_2$ is a candidate reward distribution. Conversely, every candidate reward distribution $Q$ and every coupling in $\Gamma(Q^0_{x,a},Q)$ give an element of $\mathfrak C(Q^0_{x,a})$.

Because $\cY=[0,D]$ is compact and $c$ is lower semicontinuous and bounded below, the optimal transport infimum defining $\cT_{\cY}(Q^0_{x,a},Q)$ is attained. Hence $\cT_{\cY}(Q^0_{x,a},Q)\le\epsilon_{\cY}$ is equivalent to the existence of a coupling $\gamma\in\Gamma(Q^0_{x,a},Q)$ satisfying $ \int c(y,z)\,d\gamma(y,z)\le\epsilon_{\cY}.$
Therefore, we have
\[
    q(x,a)
    =
    \inf_{\gamma\in\mathfrak C(Q^0_{x,a}):\int c(y,z)d\gamma(y,z)\le\epsilon_{\cY}}
    \int z\,d\gamma(y,z).
\]
For brevity define
\[
    A(\gamma):=\int z\,d\gamma(y,z),
    \qquad
    B(\gamma):=\int c(y,z)\,d\gamma(y,z).
\]
Then
\[
    q(x,a)=\inf_{\gamma\in\mathfrak C(Q^0_{x,a}):B(\gamma)\le\epsilon_{\cY}}A(\gamma).
\]

For $\lambda\ge0$, define the Lagrangian $ L(\gamma,\lambda) := A(\gamma)+\lambda(B(\gamma)-\epsilon_{\cY}).$
If $\gamma$ is feasible, then $B(\gamma)\le\epsilon_{\cY}$, so $  L(\gamma,\lambda)\le A(\gamma).$
For every $\lambda\ge0$, we have
\[
    \inf_{\gamma\in\mathfrak C(Q^0_{x,a})}L(\gamma,\lambda)
    \le
    \inf_{\gamma\in\mathfrak C(Q^0_{x,a}):B(\gamma)\le\epsilon_{\cY}}A(\gamma)
    =q(x,a).
\]
Taking the supremum over $\lambda\ge0$ gives weak duality:
\[
    \sup_{\lambda\ge0}\inf_{\gamma\in\mathfrak C(Q^0_{x,a})}L(\gamma,\lambda)
    \le q(x,a).
\]

We now prove equality. Since $\cY\times\cY$ is compact, $\mathfrak C(Q^0_{x,a})$ is weakly compact. The map $A$ is continuous because $z$ is bounded and continuous. The map $B$ is lower semicontinuous because $c$ is lower semicontinuous and bounded below. Since $0\le c\le C_{\cY}$, every coupling satisfies $0\le B(\gamma)\le C_{\cY}$.

Define the value function
\[
    w(t):=
    \inf_{\gamma\in\mathfrak C(Q^0_{x,a}):B(\gamma)\le t}A(\gamma),
    \qquad
    t\in[0,C_{\cY}].
\]
The diagonal coupling has cost zero because $c(y,y)=0$, so $w(t)$ is finite for all $t\in[0,C_{\cY}]$. The function $w$ is nonincreasing because a larger budget gives a larger feasible set. It is also convex: if $\gamma_1$ and $\gamma_2$ are feasible for budgets $t_1$ and $t_2$, then $\theta\gamma_1+(1-\theta)\gamma_2$ is feasible for budget $\theta t_1+(1-\theta)t_2$, and the objective value is the same convex combination. Taking near-minimizers and letting the approximation error vanish gives convexity.

If $\epsilon_{\cY}\ge C_{\cY}$, then the transport constraint is inactive. The primal value is $ \inf_{\gamma\in\mathfrak C(Q^0_{x,a})}A(\gamma),$
and this is exactly the dual value at $\lambda=0$. Now suppose $0<\epsilon_{\cY}<C_{\cY}$. Since $w$ is finite and convex on an interval, it has a subgradient at $\epsilon_{\cY}$. Let $s\in\partial w(\epsilon_{\cY})$. Since $w$ is nonincreasing, we may take $s\le0$. Set $\lambda^\star:=-s\ge0$. The subgradient inequality gives, for all $t\in[0,C_{\cY}]$,
\[
    w(t)
    \ge
    w(\epsilon_{\cY})+s(t-\epsilon_{\cY})
    =
    w(\epsilon_{\cY})-\lambda^\star(t-\epsilon_{\cY}).
\]
For any $\gamma\in\mathfrak C(Q^0_{x,a})$, set $t=B(\gamma)$. Since $A(\gamma)\ge w(B(\gamma))$,
\[
    A(\gamma)+\lambda^\star(B(\gamma)-\epsilon_{\cY})
    \ge
    w(\epsilon_{\cY}).
\]
Taking the infimum over $\gamma\in\mathfrak C(Q^0_{x,a})$ yields
\[
    \inf_{\gamma\in\mathfrak C(Q^0_{x,a})}L(\gamma,\lambda^\star)
    \ge
    w(\epsilon_{\cY})=q(x,a).
\]
Combining this with weak duality gives
\[
    q(x,a)
    =
    \sup_{\lambda\ge0}
    \inf_{\gamma\in\mathfrak C(Q^0_{x,a})}
    \left\{
        \int z\,d\gamma+
        \lambda\left(\int c(y,z)\,d\gamma(y,z)-\epsilon_{\cY}\right)
    \right\}.
\]

It remains to evaluate the inner infimum. For fixed $\lambda\ge0$, one has
\[
\begin{aligned}
    \inf_{\gamma\in\mathfrak C(Q^0_{x,a})}L(\gamma,\lambda)
    &=-\lambda\epsilon_{\cY}
    +
    \inf_{\gamma:\gamma_1=Q^0_{x,a}}
    \int\{z+\lambda c(y,z)\}\,d\gamma(y,z).
\end{aligned}
\]
Disintegrate $\gamma(dy,dz)=Q^0_{x,a}(dy)K_y(dz)$, where $K_y$ is a Markov kernel on $\cY$. Then
\[
    \int\{z+\lambda c(y,z)\}\,d\gamma(y,z)
    =
    \int
    \left[
        \int\{z+\lambda c(y,z)\}K_y(dz)
    \right]Q^0_{x,a}(dy).
\]
For each fixed $y$, every probability measure $K_y$ has average at least $\inf_{z\in\cY}\{z+\lambda c(y,z)\}.$
Therefore
\[
    \inf_{\gamma:\gamma_1=Q^0_{x,a}}
    \int\{z+\lambda c(y,z)\}\,d\gamma(y,z)
    \ge
    \E_{Y\sim Q^0_{x,a}}
    \left[
        \inf_{z\in\cY}\{z+\lambda c(Y,z)\}
    \right].
\]
Conversely, define $f_\lambda(y,z):=z+\lambda c(y,z)$. This function is lower semicontinuous, and hence Borel measurable, on the compact metric space $\cY\times\cY$. Its value function $s_\lambda(y):=\min_{z\in\cY}f_\lambda(y,z)$ is lower semicontinuous, and the argmin correspondence has nonempty compact values and a Borel graph. The Arsenin--Kunugui measurable-selection theorem therefore provides a Borel map $z_\lambda:\cY\to\cY$ satisfying
\[
    z_\lambda(y)\in\argmin_{z\in\cY}f_\lambda(y,z)
    \qquad\forall y\in\cY.
\]
Taking $K_y=\delta_{z_\lambda(y)}$ gives the reverse inequality, and hence
\[
    \inf_{\gamma:\gamma_1=Q^0_{x,a}}
    \int\{z+\lambda c(y,z)\}\,d\gamma(y,z)
    =
    \E_{Y\sim Q^0_{x,a}}
    \left[
        \inf_{z\in\cY}\{z+\lambda c(Y,z)\}
    \right].
\]
Thus
\[
    q(x,a)
    =
    \sup_{\lambda\ge0}
    \left\{
        -\lambda\epsilon_{\cY}
        +
        \E_{Y\sim Q^0_{x,a}}
        \left[
            \inf_{z\in\cY}\{z+\lambda c(Y,z)\}
        \right]
    \right\}.
\]

Finally, we restrict the dual domain. For every $y\in\cY$, we have $0\le \inf_{z\in\cY}\{z+\lambda c(y,z)\}\le D.$
The lower bound follows from $z\ge0$ and $c\ge0$. The upper bound follows by choosing $z=y$ and using $c(y,y)=0$. Hence the dual objective is at most $-\lambda\epsilon_{\cY}+D$. At $\lambda=0$, the objective equals $0$. If $\lambda>D/\epsilon_{\cY}$, then $-\lambda\epsilon_{\cY}+D<0$, so such $\lambda$ cannot improve over $\lambda=0$. Therefore the supremum may be restricted to $[0,D/\epsilon_{\cY}]$.
\end{proof}

\subsection{Reward dual integrand ordering and regularity
(Lemma~\ref{lem:reward_score_regular})}
\label{app:proof-reward-score-regular}


\begin{lemma}[Reward dual integrand ordering and regularity]
\label{lem:reward_score_regular}
For $\lambda\in[0,D/\epsilon_{\cY}]$ and $y\in\cY$, let $s_\lambda(y) := \inf_{z\in\cY}\{z+\lambda c_{\cY}(y,z)\}.$
Then:
\begin{enumerate}
    \item For every $y\in\cY$ and $0\le\lambda\le\lambda'\le D/\epsilon_{\cY}$,
    \[
        0
        \le
        s_\lambda(y)
        \le
        s_{\lambda'}(y)
        \le
        y
        \le
        D.
    \]
    \item For every $y\in\cY$ and $\lambda,\lambda'\in[0,D/\epsilon_{\cY}]$,
    \[
        |s_\lambda(y)-s_{\lambda'}(y)|
        \le
        C_{\cY}|\lambda-\lambda'|.
    \]
    \item For every $\lambda$, $y\mapsto s_\lambda(y)$ is Borel measurable, and the class $\{s_\lambda:\lambda\in[0,D/\epsilon_{\cY}]\}$ is pointwise separable by rational values of $\lambda$.
\end{enumerate}
\end{lemma}

\begin{proof}
Fix $y\in\cY$. The bounds $0\le s_\lambda(y)\le y\le D$ follow from $z\in[0,D]$, $c_{\cY}\ge0$, and the admissible choice $z=y$. If $\lambda\le\lambda'$, then for every $z\in\cY$,
\[
    z+\lambda c_{\cY}(y,z)
    \le
    z+\lambda'c_{\cY}(y,z),
\]
because the cost is nonnegative. Taking the infimum over $z$ on both sides gives $s_\lambda(y)\le s_{\lambda'}(y)$. This ordering uses no structural property of the cost beyond its standing nonnegativity assumption.

For Lipschitzness, define $f_\lambda(z):=z+\lambda c_{\cY}(y,z).$
For every $z\in\cY$, one has
\[
    |f_\lambda(z)-f_{\lambda'}(z)|
    =
    |\lambda-\lambda'|c_{\cY}(y,z)
    \le
    C_{\cY}|\lambda-\lambda'|.
\]
Thus
\[
    f_\lambda(z)
    \le
    f_{\lambda'}(z)+C_{\cY}|\lambda-\lambda'|
    \qquad \forall z\in\cY.
\]
By monotonicity of the infimum, we have
\[
\begin{aligned}
    s_\lambda(y)
    &=\inf_{z\in\cY}f_\lambda(z) \\
    &\le
    \inf_{z\in\cY}\{f_{\lambda'}(z)+C_{\cY}|\lambda-\lambda'|\} \\
    &=s_{\lambda'}(y)+C_{\cY}|\lambda-\lambda'|.
\end{aligned}
\]
Interchanging $\lambda$ and $\lambda'$ gives
\[
    |s_\lambda(y)-s_{\lambda'}(y)|
    \le
    C_{\cY}|\lambda-\lambda'|.
\]
This proves the Lipschitz inequality and, in particular, continuity in $\lambda$.

For fixed $\lambda$, the map $(y,z)\mapsto z+\lambda c_{\cY}(y,z)$ is lower semicontinuous on $\cY\times\cY$. Since the minimization is over compact $\cY$, the resulting value function $y\mapsto s_\lambda(y)$ is lower semicontinuous and hence Borel measurable. Finally, continuity in $\lambda$ implies that, for every $y$ and every $\lambda$, $s_{\lambda_k}(y)\to s_\lambda(y)$ along any rational sequence $\lambda_k\to\lambda$. Thus the countable rationally indexed subclass is pointwise dense, which supplies the claimed separability.
\end{proof}

\subsection{Context-level OT duality (Lemma~\ref{lem:context_dual})}
\label{app:proof-context-dual}

\begin{proof}
It suffices to prove the following generic identity for an arbitrary center $R\in\Delta(\cX)$ and an arbitrary real-valued function $h:\cX\to\R$.
We prove
\[
    \inf_{P\in\Delta(\cX):\cT_{\cX}(R,P)\le\epsilon_{\cX}}
    \E_{Z\sim P}[h(Z)]
    =
    \sup_{\eta\ge0}
    \left\{
        -\eta\epsilon_{\cX}
        +
        \E_{X\sim R}
        \left[\min_{z\in\cX}\{h(z)+\eta c_{\cX}(X,z)\}\right]
    \right\}.
\]
Because $\cX$ is finite, write the optimization directly in coupling variables. Let $\gamma_{xz}$ be the mass transported from $x$ to $z$. The first marginal constraint is
\[
    \sum_{z\in\cX}\gamma_{xz}=R(x),
    \qquad \forall x\in\cX.
\]
The second marginal is free. Therefore the primal problem is
\[
\begin{aligned}
    \min_{\gamma\ge0}
    &\quad
    \sum_{x,z\in\cX}\gamma_{xz}h(z)\\
    \text{subject to}
    &\quad
    \sum_{z\in\cX}\gamma_{xz}=R(x),
    \qquad \forall x\in\cX,\\
    &\quad
    \sum_{x,z\in\cX}\gamma_{xz}c_{\cX}(x,z)
    \le
    \epsilon_{\cX}.
\end{aligned}
\]
The feasible set is nonempty because the diagonal coupling $\gamma_{xx}=R(x)$ has transport cost zero.

Define
\[
    A_X(\gamma):=\sum_{x,z}\gamma_{xz}h(z),
    \qquad
    B_X(\gamma):=\sum_{x,z}\gamma_{xz}c_{\cX}(x,z).
\]
For $\eta\ge0$, define
\[
    L_X(\gamma,\eta):=A_X(\gamma)+\eta(B_X(\gamma)-\epsilon_{\cX}).
\]
For every feasible $\gamma$, we have $L_X(\gamma,\eta)\le A_X(\gamma)$, which gives weak duality.

To prove equality, define
\[
    w_X(t):=
    \min_{\gamma\ge0:\sum_z\gamma_{xz}=R(x),\ B_X(\gamma)\le t}
    A_X(\gamma),
    \qquad t\in[0,C_{\cX}].
\]
The diagonal coupling makes $w_X(t)$ finite for all $t\in[0,C_{\cX}]$. The function $w_X$ is nonincreasing and convex by the same convex-combination argument used in the reward-level proof. If $\epsilon_{\cX}\ge C_{\cX}$, the transport constraint is inactive and $\eta=0$ gives equality. Otherwise $0<\epsilon_{\cX}<C_{\cX}$; choose $s\in\partial w_X(\epsilon_{\cX})$ with $s\le0$ and set $\eta^\star=-s\ge0$. The subgradient inequality implies, for every row-coupling $\gamma$ with first marginal $R$, we have
\[
    A_X(\gamma)+\eta^\star(B_X(\gamma)-\epsilon_{\cX})
    \ge
    w_X(\epsilon_{\cX}).
\]
Taking the infimum over all row-couplings and combining with weak duality proves strong duality:
\[
    w_X(\epsilon_{\cX})
    =
    \sup_{\eta\ge0}
    \inf_{\gamma:\sum_z\gamma_{xz}=R(x)}L_X(\gamma,\eta).
\]

It remains to evaluate the inner infimum. For fixed $\eta\ge0$,
\[
\begin{aligned}
    \inf_{\gamma:\sum_z\gamma_{xz}=R(x)}L_X(\gamma,\eta)
    &=-\eta\epsilon_{\cX}
    +
    \inf_{\gamma:\sum_z\gamma_{xz}=R(x)}
    \sum_{x,z}\gamma_{xz}\{h(z)+\eta c_{\cX}(x,z)\}.
\end{aligned}
\]
The row constraints separate across $x$. For each fixed $x$, the row has mass $R(x)$ and can put all of that mass on a minimizer of $z\mapsto h(z)+\eta c_{\cX}(x,z)$. Since $\cX$ is finite, the minimum exists. Therefore
\[
    \inf_{\gamma:\sum_z\gamma_{xz}=R(x)}L_X(\gamma,\eta)
    =
    -\eta\epsilon_{\cX}
    +
    \sum_{x\in\cX}R(x)
    \min_{z\in\cX}\{h(z)+\eta c_{\cX}(x,z)\}.
\]
This is the stated dual formula.

It remains to justify the bounded dual intervals used in the manuscript.
Suppose first that $h:\cX\to[0,D]$. For every $x\in\cX$,
\[
    0
    \le \min_{z\in\cX}\{h(z)+\eta c_{\cX}(x,z)\}
    \le h(x)+\eta c_{\cX}(x,x)=h(x)\le D.
\]
Hence the dual objective is at most $-\eta\epsilon_{\cX}+D$, while its value at $\eta=0$ is $\min_z h(z)\ge0$. No $\eta>D/\epsilon_{\cX}$ can improve over $\eta=0$. Applying this with $R=P^0$ and $h(z)=\sum_a\pi_z(a)q(z,a)$ proves Lemma~\ref{lem:context_dual}.

For the empirical pessimistic value, clipping gives $h_{\pi,\underline q}:\cX\to[0,D]$. The same argument therefore shows that no $\eta>D/\epsilon_{\cX}$ can improve over $\eta=0$, proving that the interval in \eqref{eq:hatV} is valid.
\end{proof}

\subsection{Monotonicity and Lipschitzness of the context robust value (Lemma~\ref{lem:context_monotone})}
\label{app:proof-context-monotone}


\begin{lemma}[Monotonicity and Lipschitzness of the context robust value]
\label{lem:context_monotone}
For a center $R\in\Delta(\cX)$ and a function $f:\cX\to\R$, define
\[
    \Phi_R(f)
    :=
    \inf_{P:\cT_{\cX}(R,P)\le\epsilon_{\cX}}
    \E_{X\sim P}[f(X)].
\]
Let $h,g:\cX\to\R$. Then:
\begin{enumerate}
    \item If $h\le g$, then $\Phi_R(h)\le\Phi_R(g)$.
    \item $|\Phi_R(h)-\Phi_R(g)|\le\|h-g\|_\infty$.
    \item If $h\le g$, then
    \[
        0
        \le
        \Phi_R(g)-\Phi_R(h)
        \le
        \sup_{P:\cT_{\cX}(R,P)\le\epsilon_{\cX}}
        \E_{X\sim P}[g(X)-h(X)].
    \]
\end{enumerate}
\end{lemma}

\begin{proof}
Let
\[
    \mathcal U_X(R)
    :=
    \{P\in\Delta(\cX):\cT_{\cX}(R,P)\le\epsilon_{\cX}\}.
\]
This feasible set is identical for $h$ and $g$.

If $h(x)\le g(x)$ for every $x$, then for every $P\in\mathcal U_X(R)$,
\[
    \E_P[h(X)]\le \E_P[g(X)].
\]
Taking the infimum over the same feasible set gives the monotonicity inequality.

For Lipschitzness, let $r=\|h-g\|_\infty$. Then $h\le g+r$ and $g\le h+r$. By monotonicity and by the fact that adding a constant shifts every feasible expectation by that constant,
\[
    \inf_{P\in\mathcal U_X(R)}\E_P[h(X)]
    \le
    \inf_{P\in\mathcal U_X(R)}\E_P[g(X)] + r,
\]
and symmetrically,
\[
    \inf_{P\in\mathcal U_X(R)}\E_P[g(X)]
    \le
    \inf_{P\in\mathcal U_X(R)}\E_P[h(X)] + r.
\]
Combining the two inequalities gives the Lipschitz bound.

For the sharper one-sided bound, assume $h\le g$ and set $\Delta(x)=g(x)-h(x)\ge0$. Monotonicity gives the lower bound. Since $\cX$ is finite, the infimum over $\mathcal U_X(R)$ is attained. Let $P_h$ be a minimizer for $h$. Then
\[
\begin{aligned}
    \inf_{P\in\mathcal U_X(R)}\E_P[g(X)]
    -
    \inf_{P\in\mathcal U_X(R)}\E_P[h(X)]
    &\le
    \E_{P_h}[g(X)]-\E_{P_h}[h(X)]\\
    &=
    \E_{P_h}[\Delta(X)]\\
    &\le
    \sup_{P\in\mathcal U_X(R)}\E_P[\Delta(X)].
\end{aligned}
\]
This proves the sharper bound.
\end{proof}

%% file: appendix_lcb.tex
\section{Finite-Sample Confidence Bounds}
\label{app:confidence}

The confidence analysis has three parts. First, pointwise ordering of the reward dual integrands yields a cellwise bound without discretizing the dual parameter or introducing reward-support-cardinality dependence. Second, a total-variation bound controls the effect of replacing the nominal context distribution by its empirical estimate, uniformly over policies and bounded reward tables. Combining these results gives a lower confidence bound for every policy's robust value.

\subsection{Reward-cell concentration (Proposition~\ref{prop:reward_concentration})}
\label{app:proof-reward-concentration}

The key step is to control the reward dual uniformly over its continuous parameter $\lambda$ without splitting $\lambda$ into a finite grid. Pointwise ordering of the dual integrands permits a layer-cake representation and a Dvoretzky--Kiefer--Wolfowitz bound; bounded differences then turn the resulting mean bound into a high-probability bound.

\begin{proof}[Proof of Proposition~\ref{prop:reward_concentration}]
Conditional on the context--action samples
\(
    \mathcal F_{XA}:=\sigma\bigl((X_i,A_i)_{i=1}^n\bigr),
\)
every count $N_{x,a}$ is fixed. Also, within each nonempty cell $(x,a)$, the observed rewards are independent draws from $Q^0_{x,a}$.

Recall the reward dual integrand
\(
    s_\lambda(y)
    :=
    \inf_{z\in\cY}\{z+\lambda c_{\cY}(y,z)\},
\)
with $\lambda\in[0,D/\epsilon_{\cY}]$ and $y\in\cY$.
Lemma~\ref{lem:reward_score_regular} shows that these integrands are Borel measurable in $y$, bounded in $[0,D]$, continuous in $\lambda$, and nondecreasing in $\lambda$ pointwise in $y$. These are the only properties of the integrands used below.

We now prove concentration directly for these integrands. Fix a nonempty cell $(x,a)$, let $m=N_{x,a}$, and relabel its rewards as $Y_1,\ldots,Y_m$. In this subsection, write
\[
    P_m f:=\frac1m\sum_{i=1}^m f(Y_i),
    \qquad
    Pf:=\E_{Y\sim Q^0_{x,a}}[f(Y)],
    \qquad
    \Delta_m:=\sup_{0\le\lambda\le D/\epsilon_{\cY}}|(P_m-P)s_\lambda|.
\]

\textbf{Step 1: reduce the supremum to countably many values of $\lambda$.}

Let
\(
    \Lambda_{\mathbb Q}
    :=
    \left([0,D/\epsilon_{\cY}]\cap\mathbb Q\right)
    \cup\{D/\epsilon_{\cY}\}
\)
be a countable dense subset of the dual interval $[0,D/\epsilon_{\cY}]$. Pointwise continuity of $s_\lambda(y)$ implies that $P_m s_\lambda$ is continuous in $\lambda$. The bound $0\le s_\lambda\le D$ and dominated convergence imply that $P s_\lambda$ is also continuous in $\lambda$. Hence, $\Delta_m$ can be expressed as:
\[
    \Delta_m
    =
    \sup_{\lambda\in\Lambda_{\mathbb Q}}
    |(P_m-P)s_\lambda|.
\] 
Each term on the right hand side is measurable, so $\Delta_m$ is measurable.

\textbf{Step 2: bound the mean of $\Delta_m$ by layer cake and DKW.}

For every $v\in[0,D]$, we can write $v$ as a layer-cake identity:
\(
    v=\int_0^D\1\{r<v\}\,dr.
\)
Applying this identity to $s_\lambda$ yields
\[
    (P_m-P)s_\lambda
    =
    \int_0^D
    (P_m-P)\1\{r<s_\lambda\}\,dr.
\]
Taking absolute values and applying the triangle inequality, for each $\lambda\in\Lambda_{\mathbb Q}$ we have
\[
\left|(P_m-P)s_\lambda\right|
\le
\int_0^D
\left|(P_m-P)\1\{r<s_\lambda\}\right|\,dr.
\]
Taking the supremum over $\lambda$ and moving it inside the integral gives
\begin{equation}
    \Delta_m
    \le
    \int_0^D
    \sup_{\lambda\in\Lambda_{\mathbb Q}}
    \left|(P_m-P)\1\{r<s_\lambda\}\right|\,dr.
    \label{eq:reward-layer-cake}
\end{equation}

Fix $r\in[0,D]$ and enumerate $\Lambda_{\mathbb Q}$ as $\lambda_1,\lambda_2,\ldots$. Because $s_\lambda(y)$ is nondecreasing in $\lambda$, the sets $\{y:r<s_{\lambda_j}(y)\}$ are nested. We encode this family by the measurable function
\[
    H_r(y):=\sum_{j\ge1}2^{-j}\1\{r<s_{\lambda_j}(y)\}.
\]

For a fixed $j$, suppose $r<s_{\lambda_j}(y)$ but $r\ge s_{\lambda_j}(y')$. Nestedness implies that, for every $k$, we have $\1\{r<s_{\lambda_k}(y)\}\ge\1\{r<s_{\lambda_k}(y')\}$. Consequently,
\[
    H_r(y)\ge H_r(y')+2^{-j}.
\]
Hence every $y$ satisfying $r<s_{\lambda_j}(y)$ has an $H_r$ value at least $2^{-j}$ larger than every $y'$ that does not satisfy it. We can therefore choose $c_j\in\R$ between these two ranges so that
\[
    \{y:r<s_{\lambda_j}(y)\}=\{y:H_r(y)>c_j\}.
\]
If the set on the left is empty or equals $\cY$, choose $c_j$ above or below the range of $H_r$, respectively.

Conditional on $\mathcal F_{XA}$, the variables $H_r(Y_1),\ldots,H_r(Y_m)$ are i.i.d. Each event ${r<s_{\lambda_j}(Y)}$ can be written as ${H_r(Y)>c_j}$ for some $c_j$. The Dvoretzky--Kiefer--Wolfowitz--Massart inequality controls the empirical probabilities of all such thresholds simultaneously. Thus, for every $t>0$,
\[
    \Pbb\left(
        \sup_{\lambda\in\Lambda_{\mathbb Q}}
        \left|(P_m-P)\1\{r<s_\lambda\}\right|>t
        \mathrel{\Big|}\mathcal F_{XA}
    \right)
    \le 2e^{-2mt^2}.
\]

We use this tail bound to control the expected supremum at the fixed level $r$. Since the expectation of a nonnegative random variable equals the integral of its tail probabilities,
\[
    \E\left[
        \sup_{\lambda\in\Lambda_{\mathbb Q}}
        \left|(P_m-P)\1\{r<s_\lambda\}\right|
        \mathrel{\Big|}\mathcal F_{XA}
    \right]
    \le
    \int_0^\infty 2e^{-2mt^2}\,dt
    =
    \sqrt{\frac{\pi}{2m}}.
\]
This bound holds for every $r\in[0,D]$, so we can integrate it over $r$. Each term in the supremum is jointly measurable in the rewards and $r$, and the supremum is over a countable set. The integrand in~\eqref{eq:reward-layer-cake} is therefore measurable and nonnegative. Taking conditional expectations and applying Tonelli's theorem gives
\begin{equation}
    \E[\Delta_m\mid\mathcal F_{XA}]
    \le
    \int_0^D
    \E\left[
        \sup_{\lambda\in\Lambda_{\mathbb Q}}
        \left|(P_m-P)\1\{r<s_\lambda\}\right|
        \mathrel{\Big|}\mathcal F_{XA}
    \right]\,dr
    \le
    D\sqrt{\frac{\pi}{2m}}.
\label{eq:reward-mean-bound}
\end{equation}

\textbf{Step 3: turn the mean bound into a high-probability bound.}

Replace one observation $Y_i$ by another reward, and write $P_m'$ for the resulting empirical average. Since $0\le s_\lambda\le D$, for every $\lambda$,
\(
    |P_m s_\lambda-P_m's_\lambda|\le\frac{D}{m}.
\)
Let
\(
    \Delta_m':=\sup_{\lambda\in\Lambda_{\mathbb Q}}
    |(P_m'-P)s_\lambda|.
\)
The triangle inequality gives $|\Delta_m-\Delta_m'|\le D/m$.

Conditional on $\mathcal F_{XA}$, the rewards are independent. McDiarmid's inequality consequently yields
\[
    \Pbb\left(
        \Delta_m-\E[\Delta_m\mid\mathcal F_{XA}]>t
        \mathrel{\Big|}\mathcal F_{XA}
    \right)
    \le\exp\left(-\frac{2mt^2}{D^2}\right).
\]
For $u\in(0,1)$, take $t=D\sqrt{\log(1/u)/(2m)}$ and use~\eqref{eq:reward-mean-bound}. With conditional probability at least $1-u$,
\begin{equation}
    \Delta_m
    \le
    D\left\{
        \sqrt{\frac{\pi}{2m}}
        +
        \sqrt{\frac{\log(1/u)}{2m}}
    \right\}.
    \label{eq:reward-cell-uniform}
\end{equation}

\textbf{Step 4: make the reward bound simultaneous over cells.}

For a nonempty cell, the empirical and population dual objectives contain the same term $-\lambda\epsilon_{\cY}$. Their dual representations and the inequality $|\sup F-\sup G|\le\sup|F-G|$ give
\[
\begin{aligned}
    |\widehat q(x,a)-q(x,a)|
    &=
    \left|
        \sup_{\lambda\in[0,D/\epsilon_{\cY}]}
        \{P_m s_\lambda-\lambda\epsilon_{\cY}\}
        -
        \sup_{\lambda\in[0,D/\epsilon_{\cY}]}
        \{P s_\lambda-\lambda\epsilon_{\cY}\}
    \right|\\
    &\le
    \sup_{\lambda\in[0,D/\epsilon_{\cY}]}
    |(P_m-P)s_\lambda|
    =\Delta_m.
\end{aligned}
\]

Apply~\eqref{eq:reward-cell-uniform} in each nonempty cell with $m=N_{x,a}$ and $u=\delta/(2|\cX||\cA|)$. With conditional probability at least $1-\delta/(2|\cX||\cA|)$,
\[
    |\widehat q(x,a)-q(x,a)|
    \le
    D\left\{
        \sqrt{\frac{\pi}{2N_{x,a}}}
        +
        \sqrt{\frac{\log(2|\cX||\cA|/\delta)}{2N_{x,a}}}
    \right\}.
\]
Both robust rewards lie in $[0,D]$, so their difference is also at most $D$. Taking the smaller bound gives $|\widehat q(x,a)-q(x,a)|\le\beta_Y(N_{x,a};\delta)$. For an empty cell, $\widehat q(x,a)=0$ and $q(x,a)\in[0,D]$, so the same bound holds deterministically because $\beta_Y(0;\delta)=D$.

There are at most $|\cX||\cA|$ nonempty cells. A union bound gives conditional probability at least $1-\delta/2$ that the bound holds in every cell. Averaging over the context--action sample gives the same unconditional probability.

On this simultaneous event, fix any cell and write $\beta=\beta_Y(N_{x,a};\delta)$. Since $\widehat q(x,a)-\beta\le q(x,a)$ and $q(x,a)\ge0$, the clipped lower confidence bound satisfies
\[
    0\le
    \underline q(x,a)
    =\max\{0,\widehat q(x,a)-\beta\}
    \le q(x,a).
\]
Also, $\underline q(x,a)\ge\widehat q(x,a)-\beta$, so
\[
    0\le q(x,a)-\underline q(x,a)
    \le q(x,a)-\widehat q(x,a)+\beta
    \le2\beta.
\]
Finally, $\widehat q(x,a)\in[0,D]$ and $\beta\ge0$ imply $\underline q(x,a)\in[0,D]$ for every sample, with $\underline q(x,a)=0$ in an empty cell. This proves all assertions.
\end{proof}

\subsection{Total-variation context concentration (Lemma~\ref{lem:context_concentration})}
\label{app:proof-context-concentration}

\begin{lemma}[Total-variation context concentration]
\label{lem:context_concentration}
For any reward table $r:\cX\times\cA\to[0,D]$, define
\[
    V_r(\pi)
    :=
    \inf_{P:\cT_{\cX}(P^0,P)\le\epsilon_{\cX}}
    \E_{X\sim P}\left[\sum_a\pi_X(a)r(X,a)\right]
\]
and
\[
    \widehat V_r(\pi)
    :=
    \inf_{P:\cT_{\cX}(\widehat P,P)\le\epsilon_{\cX}}
    \E_{X\sim P}\left[\sum_a\pi_X(a)r(X,a)\right].
\]
Deterministically,
\[
    \left|
    \widehat V_r(\pi)-V_r(\pi)
    \right|
    \le
    \frac D2\|\widehat P-P^0\|_1
    \qquad
    \forall \pi\in\Pi,
    \quad
    \forall r:\cX\times\cA\to[0,D].
\]
With probability at least $1-\delta/2$,
\[
    \left|
    \widehat V_r(\pi)-V_r(\pi)
    \right|
    \le
    \beta_X(\delta)
    \qquad
    \forall \pi\in\Pi,
    \quad
    \forall r:\cX\times\cA\to[0,D].
\]
The high-probability event is the same for all $r$ and $\pi$. Hence both bounds also apply when the reward table and policy are chosen using the logged sample.
\end{lemma}

\begin{proof}
For every $x\in\cX$, Jensen's inequality gives
\[
    \E\left[|\widehat P(x)-P^0(x)|\right]
    \le
    \sqrt{\frac{P^0(x)(1-P^0(x))}{n}}.
\]
Consequently,
\[
    \E\left[\|\widehat P-P^0\|_1\right]
    \le
    \frac1{\sqrt n}
    \sum_{x\in\cX}\sqrt{P^0(x)(1-P^0(x))}.
\]
Replacing one context observation changes $\|\widehat P-P^0\|_1$ by at most $2/n$. McDiarmid's inequality therefore implies that, with probability at least $1-\delta/2$,
\begin{equation}
\label{eq:sharp-context-concentration}
    \|\widehat P-P^0\|_1
    \le
    \frac1{\sqrt n}
    \left[
        \sum_{x\in\cX}\sqrt{P^0(x)(1-P^0(x))}
        +\sqrt{2\log(2/\delta)}
    \right].
\end{equation}

Fix a policy $\pi\in\Pi$ and a table $r:\cX\times\cA\to[0,D]$. Define
\[
    h_{\pi,r}(z)
    :=
    \sum_{a\in\cA}\pi_z(a)r(z,a).
\]
Because $\pi_z$ is a probability distribution, we have $h_{\pi,r}(z)\in[0,D]$. The context dual may therefore be restricted to $\eta\in[0,D/\epsilon_{\cX}]$. For $\eta$ in this interval, define
\[
    g_{\eta,\pi,r}(x)
    :=
    \min_{z\in\cX}
    \left\{
        h_{\pi,r}(z)+\eta c_{\cX}(x,z)
    \right\}.
\]
Then 
\(
    0
    \le
    g_{\eta,\pi,r}(x)
    \le
    D
\) for every $x\in\cX$.
The lower bound follows from $c_{\cX}\ge0$, while the upper bound follows by choosing $z=x$ and using $c_{\cX}(x,x)=0$. Hence, for every $\eta$,
\[
\begin{aligned}
    \left|
        \E_{\widehat P}[g_{\eta,\pi,r}(X)]
        -
        \E_{P^0}[g_{\eta,\pi,r}(X)]
    \right|
    &\le
    \frac{D}{2}\|\widehat P-P^0\|_1.
\end{aligned}
\]
The dual formula for the context robust value gives
\[
\begin{aligned}
    \widehat V_r(\pi)
    &=
    \sup_{\eta\in[0,D/\epsilon_{\cX}]}
    \left\{
        -\eta\epsilon_{\cX}
        +
        \E_{\widehat P}[g_{\eta,\pi,r}(X)]
    \right\},\\
    V_r(\pi)
    &=
    \sup_{\eta\in[0,D/\epsilon_{\cX}]}
    \left\{
        -\eta\epsilon_{\cX}
        +
        \E_{P^0}[g_{\eta,\pi,r}(X)]
    \right\}.
\end{aligned}
\]
Using $|\sup f-\sup g|\le\sup|f-g|$, we obtain, without invoking any concentration event,
\[
    \left|
    \widehat V_r(\pi)-V_r(\pi)
    \right|
    \le
    \frac{D}{2}\|\widehat P-P^0\|_1.
\]
Moreover, both $\widehat V_r(\pi)$ and $V_r(\pi)$ lie in $[0,D]$, so
\[
    \left|
    \widehat V_r(\pi)-V_r(\pi)
    \right|
    \le
    D.
\]
Combining these two deterministic bounds with~\eqref{eq:sharp-context-concentration} yields, with probability at least $1-\delta/2$,
\[
    \left|
    \widehat V_r(\pi)-V_r(\pi)
    \right|
    \le
    D\min\left\{
        1,
        \frac{1}{2\sqrt n}
        \left[
            \sum_{x\in\cX}\sqrt{P^0(x)(1-P^0(x))}
            +\sqrt{2\log(2/\delta)}
        \right]
    \right\}
    =
    \beta_X(\delta).
\]
The concentration event does not depend on $r$, $\pi$, or $\eta$. Therefore, the bound holds simultaneously over all stated reward tables and policies, including data-dependent reward tables.
\end{proof}

\subsection{Policy-value lower confidence bound (Proposition~\ref{prop:value_lcb})}
\label{app:proof-value-lcb}
\begin{proof}
Let $\mathcal E_Y$ be the event of Proposition~\ref{prop:reward_concentration} and let $\mathcal E_X$ be the event of Lemma~\ref{lem:context_concentration}. They have probabilities at least $1-\delta/2$ each, so
\[
    \Pbb(\mathcal E_Y\cap\mathcal E_X)
    \ge
    1-\delta.
\]
Work on this intersection.

For every $(x,a)$, Proposition~\ref{prop:reward_concentration} gives
$0\le\underline q(x,a)\le q(x,a)$, so $\underline q:\cX\times\cA\to[0,D]$.
Therefore, for every policy $\pi$ and context $x$,
\[
    h_{\pi,\underline q}(x)
    :=
    \sum_a\pi_x(a)\underline q(x,a)
    \le
    \sum_a\pi_x(a)q(x,a)
    =:
    h_{\pi,q}(x).
\]
By Lemma~\ref{lem:context_concentration} applied to $r=\underline q$ and then by monotonicity from Lemma~\ref{lem:context_monotone},
\[
    \widehat V_{\underline q}(\pi)-\beta_X(\delta)
    \le V_{\underline q}(\pi)
    \le V_q(\pi)=V(\pi).
\]
The left-hand side is exactly $\underline V(\pi;\delta)$, hence
\(
    \underline V(\pi;\delta)\le V(\pi)
\)
for every $\pi\in\Pi$.

For the gap, write
\[
    h_\pi(x):=h_{\pi,q}(x),
    \qquad
    \underline h_\pi(x):=h_{\pi,\underline q}(x).
\]
Then
\[
\begin{aligned}
    V(\pi)-\underline V(\pi;\delta)
    &=
    V_q(\pi)-\widehat V(\pi)+\beta_X(\delta)\\
    &=
    \{V_q(\pi)-V_{\underline q}(\pi)\}
    +
    \{V_{\underline q}(\pi)-\widehat V(\pi)\}
    +
    \beta_X(\delta).
\end{aligned}
\]
Lemma~\ref{lem:context_concentration} bounds the second brace by
\[
    \beta_X(\delta).
\]
For the first brace, $\underline h_\pi\le h_\pi$, so the sharper part of Lemma~\ref{lem:context_monotone} gives
\[
\begin{aligned}
    V_q(\pi)-V_{\underline q}(\pi)
    &\le
    \sup_{P:\cT_{\cX}(P^0,P)\le\epsilon_{\cX}}
    \E_{X\sim P}[h_\pi(X)-\underline h_\pi(X)]\\
    &=
    \sup_{P:\cT_{\cX}(P^0,P)\le\epsilon_{\cX}}
    \E_{X\sim P}
    \left[
        \sum_a\pi_X(a)\{q(X,a)-\underline q(X,a)\}
    \right].
\end{aligned}
\]
Proposition~\ref{prop:reward_concentration} gives $q(x,a)-\underline q(x,a)\le2\beta_Y(N_{x,a};\delta)$ for all cells. Substituting this bound gives
\[
\begin{aligned}
    V(\pi)-\underline V(\pi;\delta)
    &\le
    2\sup_{P:\cT_{\cX}(P^0,P)\le\epsilon_{\cX}}
    \E_{X\sim P}
    \left[\sum_a\pi_X(a)\beta_Y(N_{X,a};\delta)\right]
    +2\beta_X(\delta).
\end{aligned}
\]
\end{proof}

%% file: appendix_learning.tex
\section{Policy-Learning Guarantees}
\label{app:learning}

We convert the confidence bound involving random cell counts into an explicit regret rate. A binomial concentration argument bounds each cell's reward penalty in terms of its nominal sampling probability, including cells with small expected counts. We then aggregate these bounds along an optimal robust policy using $C_{\rm rob}^\star$ and absorb the context-estimation term into the same coverage coefficient.

For $(x,a)$, define the nominal cell probability
\[
    p^0_{x,a}
    :=
    P^0(x)\pi^0_x(a).
\]

\subsection{Count-to-confidence conversion (Lemma~\ref{lem:count-to-confidence})}
\label{app:proof-count-to-confidence}
\begin{lemma}[Count-to-confidence conversion]
\label{lem:count-to-confidence}
Fix $\delta\in(0,1)$. With probability at least $1-\delta/2$, simultaneously for every $(x,a)$ with $p^0_{x,a}>0$,
\[
    \beta_Y(N_{x,a};\delta/2)
    \le
    \sqrt{\frac{2}{1-\log 2}}\,
    \frac{D}{\sqrt{np^0_{x,a}}}
    \sqrt{\log\!\left(\frac{4|\cX||\cA|}{\delta}\right)}.
\]
Moreover, on the same event,
\[
    N_{x,a}
    \ge
    \frac12 n p^0_{x,a}
    \qquad
    \text{for every $(x,a)$ such that}\qquad
    n p^0_{x,a}
    \ge
    \frac{2}{1-\log 2}
    \log\left(\frac{2|\cX||\cA|}{\delta}\right).
\]
\end{lemma}

\begin{proof}
For a fixed cell $(x,a)$, each logged sample falls in this cell with probability
\[
    p^0_{x,a}=P^0(x)\pi^0_x(a),
\]
independently across samples. Hence
\[
    N_{x,a}\sim\operatorname{Binomial}(n,p^0_{x,a}).
\]
The multiplicative Chernoff bound gives, for a binomial variable $N$ with mean $\mu$,
\[
    \Pbb(N\le(1-t)\mu)
    \le
    \left(\frac{e^{-t}}{(1-t)^{1-t}}\right)^\mu,
    \qquad 0<t<1.
\]
Taking $t=1/2$ gives
\[
    \Pbb\left(N_{x,a}<\frac12np^0_{x,a}\right)
    \le
    \exp\left(-\frac{1-\log 2}{2}np^0_{x,a}\right).
\]
For every cell satisfying the condition in the lemma,
\[
    \exp\left(-\frac{1-\log 2}{2}np^0_{x,a}\right)
    \le
    \frac{\delta}{2|\cX||\cA|}.
\]
A union bound over the at most $|\cX||\cA|$ cells shows that the displayed
count lower bound holds simultaneously with probability at least $1-\delta/2$. Work on this event and fix a cell with $p^0_{x,a}>0$.

If
\[
    n p^0_{x,a}
    \ge
    \frac{2}{1-\log 2}
    \log\left(\frac{2|\cX||\cA|}{\delta}\right),
\]
then the count lower bound and the definition of $\beta_Y$ give
\[
\begin{aligned}
    \beta_Y(N_{x,a};\delta/2)
    &\le
    \frac{D}{\sqrt{np^0_{x,a}}}
    \left\{
        \sqrt{\pi}
        +
        \sqrt{\log\!\left(\frac{4|\cX||\cA|}{\delta}\right)}
    \right\}\\
    &\le
    \sqrt{\frac{2}{1-\log 2}}\,
    \frac{D}{\sqrt{np^0_{x,a}}}
    \sqrt{\log\!\left(\frac{4|\cX||\cA|}{\delta}\right)}.
\end{aligned}
\]
The last step uses $\log(4|\cX||\cA|/\delta)\ge\log 4$ and the elementary
inequality
\[
    \sqrt{\pi}+\sqrt{u}
    \le
    \sqrt{\frac{2}{1-\log 2}}\sqrt{u},
    \qquad u\ge\log 4.
\]

If instead
\[
    n p^0_{x,a}
    <
    \frac{2}{1-\log 2}
    \log\left(\frac{2|\cX||\cA|}{\delta}\right),
\]
then the truncation $\beta_Y\le D$ gives
\[
\begin{aligned}
    \beta_Y(N_{x,a};\delta/2)
    &\le D\\
    &\le
    \sqrt{\frac{2}{1-\log 2}}\,
    \frac{D}{\sqrt{np^0_{x,a}}}
    \sqrt{\log\!\left(\frac{4|\cX||\cA|}{\delta}\right)}.
\end{aligned}
\]
Thus the count-to-confidence bound holds for every supported cell.
\end{proof}

\subsection{Regret under robust optimal-policy coverage
(Theorem~\ref{thm:coverage-rate})}
\label{app:proof-coverage-rate}
\begin{proof}
Lemma~\ref{lem:count-to-confidence} gives an event of probability at least
$1-\delta/2$.
Proposition~\ref{prop:value_lcb}, applied with confidence parameter $\delta/2$, gives another event of probability at least $1-\delta/2$. On this event, optimality of $\widehat\pi_{\delta/2}$ and the lower-confidence property imply
\[
\begin{aligned}
    V(\pi^\star)-V(\widehat\pi_{\delta/2})
    &\le
    V(\pi^\star)-\underline V(\widehat\pi_{\delta/2};\delta/2)\\
    &\le
    V(\pi^\star)-\underline V(\pi^\star;\delta/2)\\
    &\le
    2\sup_{P:\cT_{\cX}(P^0,P)\le\epsilon_{\cX}}
    \E_{X\sim P}
    \left[
        \sum_a\pi^\star_X(a)\beta_Y(N_{X,a};\delta/2)
    \right]
    +2\beta_X(\delta/2).
\end{aligned}
\]
Their intersection has probability at least $1-\delta$. Work on this
intersection. Finiteness of $C_{\rm rob}^\star$ ensures that cells with $\pi_x^\star(a)>0$ and $p^0_{x,a}=0$ have zero mass under every admissible context distribution; otherwise the coverage coefficient would be infinite.
Therefore
\[
\begin{aligned}
&\sup_{P:\cT_{\cX}(P^0,P)\le\epsilon_{\cX}}
    \E_{X\sim P}
    \left[
        \sum_a\pi^\star_X(a)\beta_Y(N_{X,a};\delta/2)
    \right]\\
&\quad\le
    \sqrt{\frac{2}{1-\log 2}}\,
    D\sqrt{\frac{\log(4|\cX||\cA|/\delta)}{n}}
    \sup_{P:\cT_{\cX}(P^0,P)\le\epsilon_{\cX}}
    \E_{X\sim P}
    \left[
        \sum_a\pi^\star_X(a)(p^0_{X,a})^{-1/2}
    \right]\\
&\quad=
    \sqrt{\frac{2}{1-\log 2}}\,
    D C_{\rm rob}^\star
    \sqrt{\frac{\log(4|\cX||\cA|/\delta)}{n}}.
\end{aligned}
\]
It remains to absorb context estimation into the same coverage term. Since $P^0$ itself belongs to the context ambiguity set and
$C_{\rm rob}^\star<\infty$,
\[
\begin{aligned}
    C_{\rm rob}^\star
    &\ge
    \sum_{x:P^0(x)>0}\sqrt{P^0(x)}
    \sum_{a\in\cA}
        \frac{\pi_x^\star(a)}{\sqrt{\pi_x^0(a)}}\\
    &\ge
    \sum_{x\in\cX}\sqrt{P^0(x)}
    \ge1.
\end{aligned}
\]
The second inequality uses $\pi_x^0(a)\le1$ on every cell used by $\pi^\star$; finiteness of the coverage coefficient excludes an uncovered such cell whenever $P^0(x)>0$. Hence
\[
    \sum_{x\in\cX}\sqrt{P^0(x)(1-P^0(x))}
    \le
    C_{\rm rob}^\star.
\]
By the definition of $\beta_X$,
\[
    2\beta_X(\delta/2)
    \le
    \frac{D}{\sqrt n}
    \left[
        \sum_{x\in\cX}\sqrt{P^0(x)(1-P^0(x))}
        +\sqrt{2\log(4/\delta)}
    \right].
\]
Since
$\log(4|\cX||\cA|/\delta)\ge\log4>1$ and
$\log(4/\delta)\le\log(4|\cX||\cA|/\delta)$, the context term is at most
\[
    (1+\sqrt2)
    D C_{\rm rob}^\star
    \sqrt{\frac{\log(4|\cX||\cA|/\delta)}{n}}.
\]
Adding this to the reward-estimation term proves the theorem.
\end{proof}

%% file: appendix_lowerbound.tex
\section{Proof of the Minimax Lower Bound (Theorem~\ref{thm:minimax-lower-bound})}
\label{app:lower-bound}

This appendix proves Theorem~\ref{thm:minimax-lower-bound}. The anchor supplies high-value context mass that the context adversary can move to a hard context, making context shift active and introducing the factor $w=1/(2r)+\rho$ into both robust regret and coverage. The hard family is indexed jointly by one of the $|\cX|-1$ hard contexts and one of the $|\cA|-1$ competing actions. A Fano argument over these $(|\cX|-1)(|\cA|-1)$ instances supplies the context--action cardinality logarithm, while a binary subfamily supplies the confidence dependence.

\subsection{Step 1: Hard instances}
\label{app:lower-bound-hard-instances}

Let
\[
    \sigma:=\frac{\epsilon_{\cY}}{D^2}\in(0,1/2],
    \qquad
    r:=|\cX|-1,
    \qquad
    L:=|\cA|-1,
    \qquad
    \rho:=\frac{\epsilon_{\cX}}{\kappa}.
\]
Label the context and action sets as
\[
    \cX=\{x_1,\ldots,x_r,x_{\rm anc}\},
    \qquad
    \cA=\{0,1,\ldots,L\},
\]
where $x_{\rm anc}$ is the anchor context. The nominal context distribution and context transport cost are
\[
    P^0(x_t)=\frac{1}{2r},
    \quad t\in\{1,\ldots,r\},
    \qquad
    P^0(x_{\rm anc})=\frac12,
    \qquad
    c_{\cX}(x,z)=\kappa\1\{x\ne z\}.
\]
Write
\[
    b:=1-L\mu.
\]
At each hard context, the behavior policy chooses every competing action with probability $\mu$ and action $0$ with probability $b$:
\[
    \pi^0_{x_t}(a)=\mu,
    \quad a\in\{1,\ldots,L\},
    \qquad
    \pi^0_{x_t}(0)=b.
\]
At the anchor it always chooses action $0$. The condition
\[
    \mu\le
    \frac{w^2}{1+Lw^2},
    \qquad
    w:=\frac{1}{2r}+\rho,
\]
implies
\begin{equation}
\label{eq:lower-bound-mu-comparison}
    \mu\le w^2b<b,
\end{equation}
so the behavior policy is well defined.

The policy class is
\[
    \Pi
    =
    \{\pi_{t,j}:t\in\{1,\ldots,r\},\ j\in\{1,\ldots,L\}\},
\]
where
\[
    \pi_{t,j}(x_t)=j,
    \qquad
    \pi_{t,j}(x)=0
    \quad\text{for }x\ne x_t.
\]
Thus every policy chooses one competing action at one hard context and chooses action $0$ everywhere else.

For $\theta\in[0,1]$, let
\[
    Q_\theta=(1-\theta)\delta_0+\theta\delta_D,
    \qquad
    \theta_0:=\frac{1+\sigma}{2}.
\]
Fix
\[
    0<h\le\frac{1-\sigma}{4}.
\]
For every $(t,j)\in\{1,\ldots,r\}\times\{1,\ldots,L\}$, define an instance $\mathsf M_{t,j}$ as follows. At every hard context, action $0$ has the common reward law
\[
    Q^0_{x_s,0}=Q_{\theta_0},
    \qquad s\in\{1,\ldots,r\}.
\]
The single elevated competing cell has reward law
\[
    Q^0_{x_t,j}=Q_{\theta_0+h},
\]
and every other competing cell has reward law
\[
    Q^0_{x_s,a}=Q_{\theta_0-h},
    \qquad
    (s,a)\ne(t,j),
    \quad a\in\{1,\ldots,L\}.
\]
At the anchor, set $Q^0_{x_{\rm anc},0}=Q_1$; reward laws of unused anchor actions may be set to $Q_0$. The instances have the same context law, behavior policy, policy class, costs, and ambiguity radii. They differ only in which of the $rL$ competing cells is elevated.

\subsection{Step 2: Robust regret and coverage}
\label{app:lower-bound-regret-coverage}

For any coupling $(Y,Z)$ with $Y\sim Q_\theta$, $Z\in[0,D]$, and $\E[(Y-Z)^2]\le\epsilon_{\cY}$, nonnegativity of $Z$ and Cauchy--Schwarz give
\[
\begin{aligned}
    D\theta-\E[Z]
    &=\E[Y-Z]\\
    &\le
    \E[(Y-Z)\1\{Y=D\}]\\
    &\le
    \sqrt{\theta\epsilon_{\cY}}.
\end{aligned}
\]
When $\theta>\sigma$, equality is attained by coupling $Z=0$ when $Y=0$ and $Z=D-\sqrt{\epsilon_{\cY}/\theta}$ when $Y=D$. Consequently,
\begin{equation}
\label{eq:lower-bound-robust-bernoulli-mean}
    q(\theta)
    :=
    \inf_{Q:W_2^2(Q_\theta,Q)\le\epsilon_{\cY}}
        \E_{Y\sim Q}[Y]
    =
    D\bigl(\theta-\sqrt{\sigma\theta}\bigr).
\end{equation}
The restriction on $h$ places $\theta_0-h,\theta_0,\theta_0+h$ in
\[
    \left[\frac{1+3\sigma}{4},\frac{3+\sigma}{4}\right]
    \subset(\sigma,1).
\]
On this interval,
\[
    q'(\theta)
    =
    D\left(1-\frac{\sqrt\sigma}{2\sqrt\theta}\right)
    \ge\frac D2.
\]
Writing
\[
    q_-:=q(\theta_0-h),
    \qquad
    q_0:=q(\theta_0),
    \qquad
    q_+:=q(\theta_0+h),
    \qquad
    q_{\rm s}:=q(1),
\]
the mean-value theorem gives
\begin{equation}
\label{eq:lower-bound-reward-gaps}
    q_+-q_0\ge\frac{Dh}{2},
    \qquad
    q_0-q_-\ge\frac{Dh}{2}.
\end{equation}
Moreover, $q_{\rm s}>q_+>q_0>q_-$.

For the discrete context cost, the diagonal mass of any coupling is at most $\sum_x\min\{P^0(x),P(x)\}$, and this bound is attainable by coupling the remaining excess and deficit mass off the diagonal. Thus
\[
\begin{aligned}
    \cT_{\cX}(P^0,P)
    &=
    \kappa\left(1-\sum_{x\in\cX}\min\{P^0(x),P(x)\}\right)\\
    &=
    \kappa\lVert P-P^0\rVert_{\rm TV}.
\end{aligned}
\]
For a linear objective, the positive and negative parts of $P-P^0$ have the same mass, equal to $\lVert P-P^0\rVert_{\rm TV}$. Hence each unit of mass decreases the objective most by moving it from a largest coefficient to a smallest coefficient (and increases it most by reversing this move). Here the anchor contains $1/2>\rho$ mass and every hard context can receive $\rho<1/2$ mass, so moving the full TV budget in this way is feasible. Under $\mathsf M_{t,j}$, the policy $\pi_{t,j}$ has robust reward $q_+$ at $x_t$, $q_0$ at the other hard contexts, and $q_{\rm s}$ at the anchor. Since $r\ge2$, the context adversary moves $\rho$ mass from the anchor to a hard context with reward $q_0$. Therefore
\[
    V_{t,j}(\pi_{t,j})
    =
    \frac{q_+}{2r}
    +
    \left(\frac12-\frac{1}{2r}+\rho\right)q_0
    +
    \left(\frac12-\rho\right)q_{\rm s}.
\]
Every other policy $\pi_{s,k}\ne\pi_{t,j}$ receives reward $q_-$ at its selected competing cell and $q_0$ at every other hard context. Its context adversary moves $\rho$ anchor mass to that low-reward context, so
\[
    V_{t,j}(\pi_{s,k})
    =
    \left(\frac{1}{2r}+\rho\right)q_-
    +
    \left(\frac12-\frac{1}{2r}\right)q_0
    +
    \left(\frac12-\rho\right)q_{\rm s}.
\]
It follows from~\eqref{eq:lower-bound-reward-gaps} that $\pi_{t,j}$ is uniquely optimal and every wrong policy incurs robust regret at least
\begin{equation}
\label{eq:lower-bound-policy-gap}
\begin{aligned}
    V_{t,j}(\pi_{t,j})-V_{t,j}(\pi_{s,k})
    &=
    \frac{q_+-q_-}{2r}
    +\rho(q_0-q_-)\\
    &\ge
    \frac{Dh}{2r}+\frac{\rho Dh}{2}\\
    &=
    \frac{wDh}{2}+\frac{Dh}{4r}\\
    &\ge
    \frac{wDh}{2}.
\end{aligned}
\end{equation}
Both shifts are active: the context adversary strictly lowers the optimal policy value by moving $\rho$ mass away from the anchor, and \eqref{eq:lower-bound-robust-bernoulli-mean} is strictly smaller than the nominal mean for every relevant reward law.

For the optimal policy $\pi_{t,j}$, the coverage weights of the selected context--action cells are
\[
    \sqrt{\frac{2r}{\mu}}
    \quad\text{at }x_t,
    \qquad
    \sqrt{\frac{2r}{b}}
    \quad\text{at every other hard context},
    \qquad
    \sqrt2
    \quad\text{at the anchor}.
\]
By~\eqref{eq:lower-bound-mu-comparison}, the first weight is the largest and the anchor weight is the smallest. The coverage adversary consequently moves $\rho$ mass from the anchor to $x_t$, giving every instance the common coefficient
\begin{equation}
\label{eq:lower-bound-coverage}
    C_{\rm rob}^\star
    =
    w\sqrt{\frac{2r}{\mu}}
    +
    \left(\frac12-\frac{1}{2r}\right)
        \sqrt{\frac{2r}{b}}
    +
    \left(\frac12-\rho\right)\sqrt2.
\end{equation}
Let $H:=\sqrt{2r/\mu}$. Since $\sqrt{\mu/b}\le w$,
\[
    \left(\frac12-\frac{1}{2r}\right)
        \sqrt{\frac{2r}{b}}
    \le
    \frac{wH}{2}.
\]
Also, $\sqrt\mu\le w\sqrt b\le w$, and hence
\[
    \left(\frac12-\rho\right)\sqrt2
    \le
    \frac{\sqrt\mu}{2\sqrt r}H
    \le
    \frac{wH}{2}.
\]
Thus
\begin{equation}
\label{eq:lower-bound-coverage-comparison}
    C_{\rm rob}^\star
    \le
    2w\sqrt{\frac{2r}{\mu}},
    \qquad\text{and hence}\qquad
    w\sqrt{\frac r\mu}
    \ge
    \frac{C_{\rm rob}^\star}{2\sqrt2}.
\end{equation}

\subsection{Step 3: Relative entropy}
\label{app:lower-bound-relative-entropy}

Let $\Pbb_{t,j}^n$ be the logged-data law under $\mathsf M_{t,j}$. Any two distinct instances differ only at their two elevated-cell indices. Each such cell has nominal probability $\mu/(2r)$. The chain rule for relative entropy therefore gives
\[
    \operatorname{KL}
        (\Pbb_{t,j}^n\Vert\Pbb_{s,k}^n)
    =
    \frac{n\mu}{2r}\bigg\{
        \operatorname{kl}(\theta_0+h,\theta_0-h)
        +
        \operatorname{kl}(\theta_0-h,\theta_0+h)
    \bigg\}.
\]
For Bernoulli distributions,
\[
    \operatorname{kl}(u,v)
    \le
    \frac{(u-v)^2}{v(1-v)}.
\]
Every parameter belongs to
$[(1+3\sigma)/4,(3+\sigma)/4]$, on which
\[
    v(1-v)
    \ge
    \frac{(3+\sigma)(1-\sigma)}{16}.
\]
Since the parameters differ by $2h$,
\begin{equation}
\label{eq:lower-bound-kl}
    \operatorname{KL}
        (\Pbb_{t,j}^n\Vert\Pbb_{s,k}^n)
    \le
    \frac{64n\mu h^2}
    {r(3+\sigma)(1-\sigma)}.
\end{equation}

\subsection{Step 4: Fano and binary testing}
\label{app:lower-bound-testing}

Let
\[
    K:=rL=(|\cX|-1)(|\cA|-1).
\]
Because $|\cX|,|\cA|\ge3$, we have $K\ge4$. We treat separately the cardinality-dominated and confidence-dominated regimes. Randomized learners are included by adjoining their independent random seed to the logged data, which does not change any relative entropy.

Suppose first that
\[
    \log K\ge\log(1/\delta).
\]
Set
\[
    h
    :=
    \sqrt{
        \frac{(3+\sigma)(1-\sigma)r\log K}
        {512n\mu}
    }.
\]
Since
\[
    \frac{3+\sigma}{1-\sigma}\le7,
    \qquad
    \log K
    \le
    \log\left(\frac{4|\cX||\cA|}{\delta}\right),
\]
the sample-size condition in the theorem implies $h\le(1-\sigma)/4$. Equation~\eqref{eq:lower-bound-kl} gives
\[
    \max_{(t,j)\ne(s,k)}
    \operatorname{KL}
        (\Pbb_{t,j}^n\Vert\Pbb_{s,k}^n)
    \le
    \frac18\log K.
\]
Fano's inequality under the uniform prior on the $K$ instances yields
\[
\begin{aligned}
    \frac1K
    \sum_{t=1}^r\sum_{j=1}^L
    \Pbb_{t,j}^n(\widehat\pi\ne\pi_{t,j})
    &\ge
    1-\frac18-\frac{\log2}{\log K}\\
    &\ge\frac38.
\end{aligned}
\]
Hence, in at least one instance, the learner chooses a wrong policy with probability at least $3/8>\delta$. On this event, \eqref{eq:lower-bound-policy-gap} and \eqref{eq:lower-bound-coverage-comparison} give
\[
\begin{aligned}
    V_{t,j}(\pi_{t,j})-V_{t,j}(\widehat\pi)
    &\ge
    \frac{D\sqrt{(3+\sigma)(1-\sigma)}}{128}
    C_{\rm rob}^\star
    \sqrt{\frac{\log K}{n}}\\
    &\ge
    \frac{\sqrt7}{256}
    D C_{\rm rob}^\star
    \sqrt{\frac{\log K}{n}},
\end{aligned}
\]
where the last inequality uses $\sigma\le1/2$.

Suppose now that
\[
    \log K<\log(1/\delta).
\]
Use any two distinct instances and set
\[
    h
    :=
    \sqrt{
        \frac{(3+\sigma)(1-\sigma)r
        \operatorname{kl}(1-\delta,\delta)}
        {128n\mu}
    }.
\]
For $\delta\in(0,1/4]$,
\[
    \frac14\log\frac1\delta
    \le
    \operatorname{kl}(1-\delta,\delta)
    \le
    \log\frac1\delta.
\]
The theorem's sample-size condition, the upper bound above, and $\log(1/\delta)\le\log(4|\cX||\cA|/\delta)$ imply $h\le(1-\sigma)/4$, while \eqref{eq:lower-bound-kl} gives
\[
    \operatorname{KL}
        (\Pbb_{t,j}^n\Vert\Pbb_{s,k}^n)
    \le
    \frac12\operatorname{kl}(1-\delta,\delta).
\]
If the learner chose the correct policy with probability at least $1-\delta$ in both instances, data processing applied to $\1\{\widehat\pi=\pi_{t,j}\}$ would give
\[
    \operatorname{KL}
        (\Pbb_{t,j}^n\Vert\Pbb_{s,k}^n)
    \ge
    \operatorname{kl}(1-\delta,\delta),
\]
a contradiction. Thus, in at least one of the two instances, the learner chooses a wrong policy with probability greater than $\delta$. On this event,
\[
\begin{aligned}
    V_{t,j}(\pi_{t,j})-V_{t,j}(\widehat\pi)
    &\ge
    \frac{D\sqrt{(3+\sigma)(1-\sigma)}}{64}
    C_{\rm rob}^\star
    \sqrt{
        \frac{\operatorname{kl}(1-\delta,\delta)}{n}
    }\\
    &\ge
    \frac{\sqrt7}{256}
    D C_{\rm rob}^\star
    \sqrt{\frac{\log(1/\delta)}{n}}.
\end{aligned}
\]

In either regime, some instance has error probability greater than $\delta$ and regret on the error event at least
\[
    \frac{\sqrt7}{256}
    D C_{\rm rob}^\star
    \sqrt{
        \frac{
            \max\{\log K,\log(1/\delta)\}
        }{n}
    }.
\]
Since $r,L\ge2$,
\[
    |\cX|=r+1\le\frac32r,
    \qquad
    |\cA|=L+1\le\frac32L.
\]
Consequently,
\[
    4|\cX||\cA|
    \le9K
    \le K^3,
\]
where the last inequality uses $K\ge4$. Hence
\[
    \log\left(\frac{4|\cX||\cA|}{\delta}\right)
    \le
    4\max\left\{
        \log K,
        \log\frac1\delta
    \right\}.
\]
Substitution proves Theorem~\ref{thm:minimax-lower-bound}.

Finally, taking $\delta=1/4$ gives an error probability greater than $1/4$. Since regret is nonnegative, the same construction gives, whenever
\[
    n\mu
    \ge
    \frac{7r}{8}\log(16|\cX||\cA|),
\]
an instance satisfying
\[
    \E\left[V(\pi^\star)-V(\widehat\pi)\right]
    \ge
    \frac{\sqrt7}{2048}
    D C_{\rm rob}^\star
    \sqrt{\frac{\log(16|\cX||\cA|)}{n}}.
\]

%% file: appendix_experiments.tex
\section{Experiment Details}
\label{app:experiments}

\subsection{Empirical Behavior of the Robust-Reward Confidence Radius}
\label{app:experiment-q-confidence}

We examine the finite-sample confidence radius in Proposition~\ref{prop:reward_concentration} across a collection of random and structured reward distributions. The experiment isolates reward estimation within a single context--action cell, setting $|\cX||\cA|=1$, and compares the largest empirical error quantile across the tested distributions with the theoretical radius.

\paragraph{Reward distributions and ambiguity set.}
We take $\cY=[0,1]$, so $D=1$, and use the quadratic transport cost $c_{\cY}(y,z)=|y-z|^2$ with ambiguity radius $\epsilon_{\cY}=0.04$. We evaluate $J=58$ nominal reward distributions: $50$ randomly generated distributions and $8$ structured cases. For each random instance, we first draw its support size uniformly from $\{2,\ldots,99\}$. Conditional on this size, we draw the support points independently from the uniform distribution on $[0,1]$ and their probability vector from $\operatorname{Dirichlet}(1,\ldots,1)$.

The structured cases broaden the test to include balanced and highly unbalanced probabilities, concentrated supports, and degenerate rewards:
\[
\begin{array}{lll}
\text{Instance} & \text{Support} & \text{Probabilities}\\
\hline
\text{Balanced Bernoulli}
    & (0,1) & (0.5,0.5)\\
\text{Rare high reward}
    & (0,1) & (0.99,0.01)\\
\text{Rare low reward}
    & (0,1) & (0.01,0.99)\\
\text{Support near zero}
    & (0,0.01,0.05) & (0.1,0.2,0.7)\\
\text{Support near one}
    & (0.95,0.99,1) & (0.2,0.3,0.5)\\
\text{Near-zero robust reward}
    & (0,1) & (0.959,0.041)\\
\text{Deterministic zero}
    & \{0\} & (1)\\
\text{Deterministic one}
    & \{1\} & (1)
\end{array}
\]
Although the nominal distributions have finite supports, the adversary may transport rewards anywhere in $[0,1]$.

\paragraph{Computation of robust rewards.}
For instance $j$, let $Q_j^0$ denote its nominal distribution and $q_j$ its true robust reward. We compute both $q_j$ and its empirical estimate using the continuous quadratic-cost problem. For any nominal distribution $Q$, if $\E_Q[Y^2]\le\epsilon_{\cY}$, transporting all rewards to zero is feasible and the robust reward equals zero. Otherwise, the optimal transport maps
\[
    Y\longmapsto \max\{0,Y-t_Q\},
\]
where $t_Q$ satisfies
\[
    \E_Q\!\left[\min\{Y,t_Q\}^2\right]
    =
    \epsilon_{\cY}.
\]
The robust reward is therefore $\E_Q[\max\{0,Y-t_Q\}]$. We solve the scalar equation by bisection, using the known probabilities for $q_j$ and the empirical frequencies for $\widehat q_{j,m}$. This computation is equivalent to the reward dual formulation and requires no discretization of the adversarial reward space. As a numerical check, we compare the batched implementation with an independent scalar dual optimizer on $232$ population and empirical distributions; the maximum absolute discrepancy is $1.28\times10^{-14}$.

\paragraph{Monte Carlo design.}
We use $20$ geometrically spaced sample sizes between $10$ and $10^6$, rounded to integers. For every instance--sample-size pair, we generate $R=1000$ independent replications, giving $58\times20\times1000=1{,}160{,}000$ replications in total. For computational efficiency, we sample multinomial reward counts directly. These counts have exactly the same distribution as those obtained from $m$ independent draws from the corresponding nominal distribution.

We set $\delta=0.1$. With $D=1$ and $|\cX||\cA|=1$, the theoretical confidence radius is
\[
    \beta_Y(m;0.1)
    =
    \min\left\{
        1,\,
        \sqrt{\frac{\pi}{2m}}
        +
        \sqrt{\frac{\log 20}{2m}}
    \right\}.
\]
Proposition~\ref{prop:reward_concentration} gives, for each fixed instance and sample size,
\[
    \Pr\!\left(
        |\widehat q_{j,m}-q_j|
        \le \beta_Y(m;0.1)
    \right)
    \ge 1-\frac{\delta}{2}
    =0.95.
\]
We therefore compute the empirical $0.95$ absolute-error quantile separately for each instance,
\[
    \widehat e_{j,0.95}(m)
    :=
    \operatorname{Quantile}_{0.95}
    \left(
        \left\{
            |\widehat q_{j,m}^{(r)}-q_j|
        \right\}_{r=1}^{R}
    \right),
\]
and plot
\[
    \widehat e_{\max,0.95}(m)
    :=
    \max_{1\le j\le J}\widehat e_{j,0.95}(m)
\]
against $\beta_Y(m;0.1)$. Taking the maximum prevents averaging across instances from obscuring a distribution with a larger error quantile. The $58$ instances are separate experiments, rather than $58$ context--action cells, so their number does not enter the cell-count factor in $\beta_Y$.

\begin{figure}[t]
    \centering
    \includegraphics[width=0.7\textwidth]
        {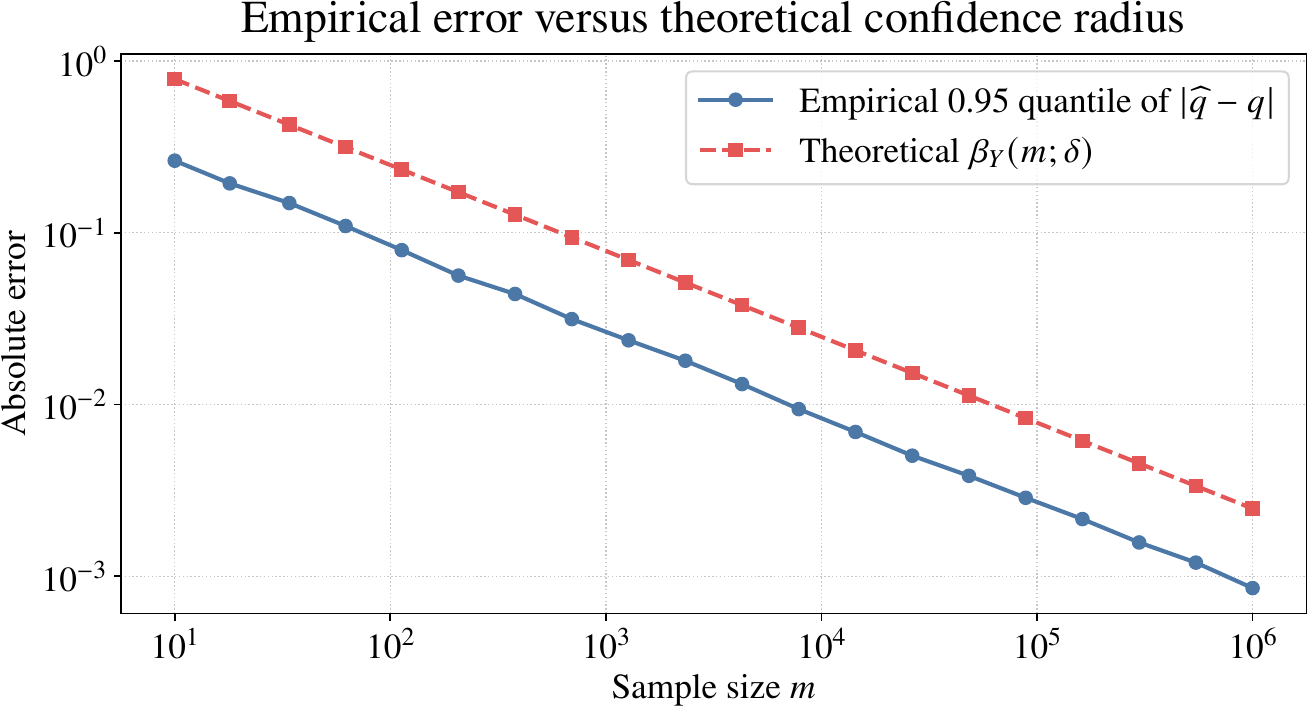}
    \caption{
        Empirical error versus theoretical confidence radius. At each sample size, the solid curve shows the maximum across $58$ nominal reward distributions of the empirical $0.95$ quantile of $|\widehat q_{j,m}-q_j|$, computed from $1000$ replications per distribution. The dashed curve shows $\beta_Y(m;0.1)$. Both axes use logarithmic scales.
    }
    \label{fig:q-confidence-bound}
\end{figure}

\paragraph{Results.}
Figure~\ref{fig:q-confidence-bound} shows that the largest empirical $0.95$ error quantile remains below the theoretical radius at every tested sample size. Across all $1160$ instance--sample-size pairs,
\[
    \max_{j,m}
    \frac{\widehat e_{j,0.95}(m)}
         {\beta_Y(m;0.1)}
    =
    0.358228.
\]
Moreover, no individual replication violates $|\widehat q_{j,m}-q_j|\le\beta_Y(m;0.1)$. The balanced Bernoulli distribution attains the largest empirical error quantile among the tested instances at every sample size. At $m=10$, this quantile is approximately $0.262536$, compared with a theoretical radius of $0.783355$. At $m=10^6$, the corresponding values are approximately $0.000854$ and $0.002477$.

The two curves exhibit similar decay on logarithmic axes. A log--log regression using the reported empirical quantiles gives a slope of approximately $-0.498$, close to the $m^{-1/2}$ dependence of the theoretical radius.

\paragraph{Interpretation.}
These results provide numerical evidence consistent with the confidence guarantee across the tested random and structured distributions, while illustrating its $m^{-1/2}$ dependence. The theoretical radius is at least approximately $2.79$ times the empirical $0.95$ error quantile for every tested pair, indicating conservative constants in this experiment. Such conservatism is compatible with a distribution-free bound that controls the reward dual integrands uniformly.

The experiment does not establish universal validity or worst-case tightness. The reported maximum is taken over a finite collection of empirical quantiles and is subject to Monte Carlo variability. In particular, it is not the $0.95$ quantile of a joint maximum error, nor does the plot assert simultaneous $95\%$ coverage across all instances and sample sizes. The varying nominal support sizes also provide a check across different discrete distributions, but this experiment does not isolate the effect of support cardinality or establish the same behavior for every transport cost and ambiguity radius.

\subsection{Dependence of Regret on $n$ and $C^*_{\rm rob}$}
\label{app:experiment-regret-n-c}

We examine the dependence of Algorithm~\ref{alg:opl}'s robust regret on the sample size $n$ and the robust optimal-policy coverage coefficient $C^\star_{\rm rob}$. With the remaining parameters fixed, the upper bound in Theorem~\ref{thm:coverage-rate} scales as $C^\star_{\rm rob}/\sqrt{n}$. Our experiment examines when this scaling is observed over a controlled family of instances.

\paragraph{Instance construction.}
We use $|\cX|=4$ uniformly distributed contexts, $|\cA|=9$ actions indexed by $0,\ldots,8$, and reward range $\cY=[0,D]$ with $D=1$. The context ambiguity set is the total-variation ball $\{P:\|P-P^0\|_{\rm TV}\le\rho\}$ with $\rho=0.2$. For rewards, we use squared transport cost $c_{\cY}(y,z)=(y-z)^2$ and radius $\epsilon_{\cY}=0.05 D^2=0.05$. A nominal reward $Y\sim\operatorname{Bernoulli}(\theta)$, has robust mean
\[
    q(\theta)=D\max\{\theta-\sqrt{\epsilon_\cY\theta},0\}.
\]
At every context, action $0$ has deterministic reward zero, actions $2,\ldots,8$ have Bernoulli parameter $\theta_0=0.65$, and action $1$ has parameter $\theta_1$ chosen so that $q(\theta_1)=q(\theta_0)+\Delta$, for a positive robust reward gap $\Delta$. Thus, action $1$ is optimal at every context and $V(\pi^\star)=q(\theta_1)$.

To prescribe a coverage coefficient $C$, we set
\[
    \mu=\frac{|\cX|}{C^2},\qquad
    \pi_x^0(a)=\mu\quad(a=1,\ldots,8),\qquad
    \pi_x^0(0)=1-8\mu.
\]
These probabilities are valid throughout the sweep. Since $P^0(x)\pi_x^0(1)=C^{-2}$ at every context, the coverage coefficient is exactly
\[
    C^\star_{\rm rob}
    =\sup_{P:\|P-P^0\|_{\rm TV}\le\rho}
      \sum_x\frac{P(x)}{\sqrt{P^0(x)\pi_x^0(1)}}
    =C.
\]
Increasing $C$ therefore reduces the expected number of observations in each optimal-action cell to $n/C^2$, while preserving the context distribution, action set, and ambiguity radii.

\paragraph{Learner and regret evaluation.}
We fix $\delta=0.1$ and use the confidence split from Theorem~\ref{thm:coverage-rate}:
\[
    \widehat\pi_{\delta/2}(x)\in
    \arg\max_{a\in\cA}
    \{\widehat q(x,a)-\beta_Y(N_{x,a};\delta/2)\}.
\]
Here $\widehat q(x,a)=q(\widehat\theta_{x,a})$ is computed from the empirical Bernoulli frequency. The confidence penalty has multiplier one, and the pessimistic scores are not clipped. Empty cells receive score $-D$, and exact ties are resolved by the smallest action index. For each learned policy, we compute its population robust regret as
\[
    R= q(\theta_1)-
    \min_{P:\|P-P^0\|_{\rm TV}\le\rho}
       \sum_x P(x)q\bigl(x,\widehat\pi_{\delta/2}(x)\bigr).
\]
The minimization uses the known instance parameters and is solved exactly; there is no additional Monte Carlo error from policy-value evaluation. Logged datasets are simulated through multinomial cell counts and conditional binomial success counts. These are exact sufficient statistics for $n$ independent logged observations and allow large sample sizes without storing every observation.

\paragraph{Sweep and independent gap selection.}
We evaluate all $425$ combinations of $25$ geometrically spaced sample sizes from $2^{10}$ to $2^{30}$, rounded to integers, and $17$ geometrically spaced coverage coefficients from $16$ to $128$. At each combination, we search the same fixed grid of $61$ geometrically spaced gaps in $[10^{-5}D,0.18D]$. Using $1{,}000$ training-dataset repetitions per candidate gap, we select the gap with the largest estimated mean regret. We then estimate regret at this selected gap using $5{,}000$ fresh, independent repetitions, giving $2{,}125{,}000$ evaluation repetitions in total. Within each search, cell counts and decoy-action observations are shared across candidate gaps; each candidate retains the correct logged-data marginal distribution.

This procedure approximates the maximum expected regret over the gap grid, rather than following a single fixed reward distribution as $n$ changes. The target exponents do not enter the gap grid or selection rule, and independent evaluation avoids reusing the selection data to report regret. The upper gap endpoint is selected in $111$ of the $425$ configurations; the lower endpoint is never selected. Consequently, the small-sample results also reflect the finite gap range, and should not be interpreted as the exact worst case over all reward distributions. Error bars are the evaluation mean plus or minus $1.96$ Monte Carlo standard errors, conditional on the selected gap; they do not quantify uncertainty in identifying the maximizing gap.

\paragraph{Small-sample behavior.}
The relevant expected cell count is $m=n/(C^\star_{\rm rob})^2$. At $n=2^{10}$, it ranges from $4$ for $C^\star_{\rm rob}=16$ to $0.0625$ for $C^\star_{\rm rob}=128$. In the latter case, an optimal-action cell is empty with probability $(1-128^{-2})^{1024}\approx0.9394$. For our fixed confidence level and problem dimensions, the penalty is
\[
    \beta_Y(N;\delta/2)
    =D\min\left\{1,\frac{k}{\sqrt N}\right\},\qquad
    k=\sqrt{\frac{\pi}{2}}+
      \sqrt{\frac{1}{2}\log\frac{4|\cX||\cA|}{\delta}}
      \approx3.1602,
    \qquad N\ge1.
\]
Thus, the penalty saturates at $D$ for counts $N\le9$. Poorly observed actions can have lower pessimistic scores than the frequently observed zero-reward action, resulting in substantial regret. Increasing $n$ eventually reduces the penalties enough for the learner to select among the positive-reward actions. This transition need not follow a single power law.

Boundedness also prevents an exact $C^\star_{\rm rob}/\sqrt n$ relationship throughout the full range. On the event of the theoretical guarantee, bounded regret and Theorem~\ref{thm:coverage-rate} together imply a bound of the form
\[
    R\le\min\left\{
        D,\;K D C^\star_{\rm rob}
        \sqrt{\frac{\log(4|\cX||\cA|/\delta)}{n}}
    \right\},
\]
where $K$ is a theorem-dependent constant. This is a high-probability upper bound, not an equality for the mean regret plotted here.

\begin{figure}[ht]
    \centering
    \includegraphics[width=0.7\textwidth]{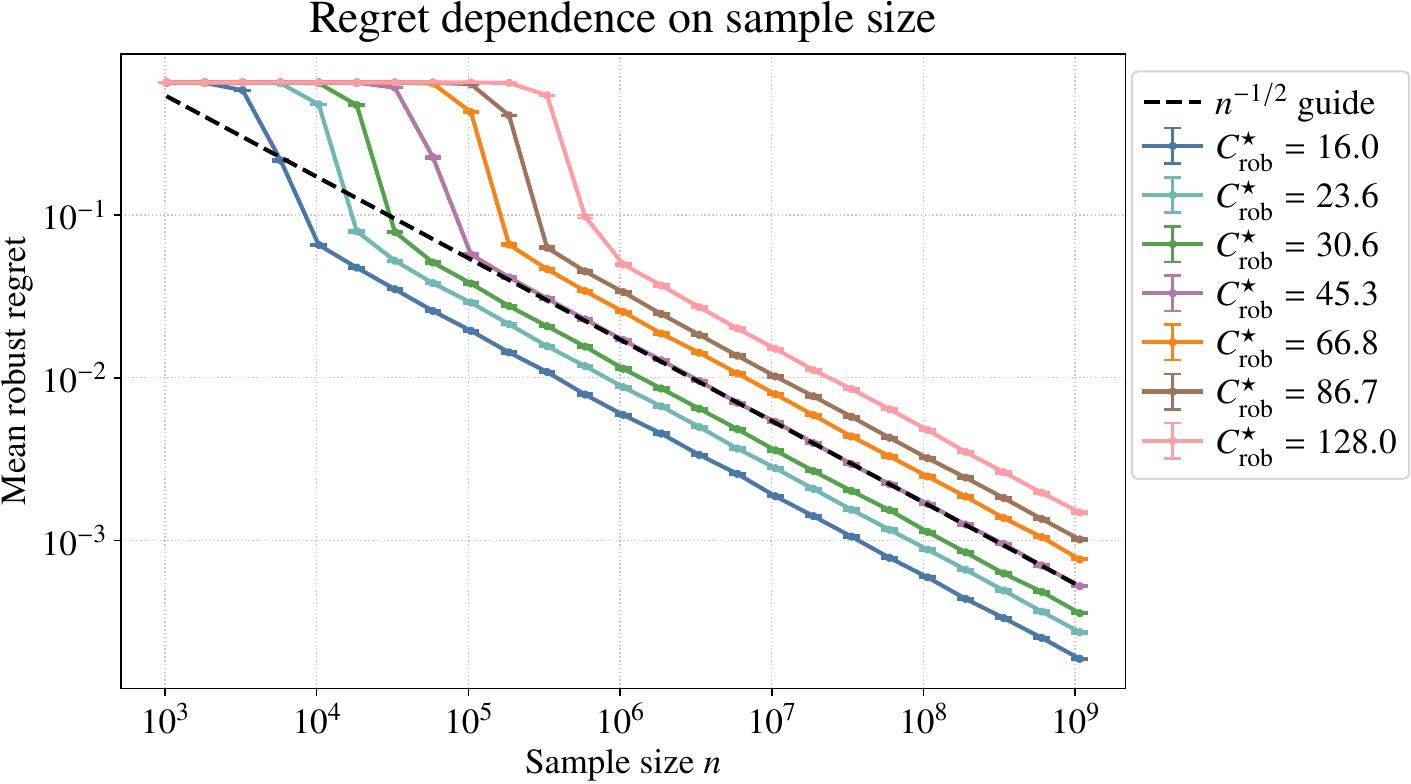}
    \caption{Mean population robust regret versus sample size for seven coverage coefficients. Each point uses independent evaluation data after gap selection. Error bars show $1.96$ Monte Carlo standard errors; the dashed line is an $n^{-1/2}$ scaling guide.}
    \label{fig:regret-vs-n}
\end{figure}

\begin{figure}[ht]
    \centering
    \includegraphics[width=0.7\textwidth]{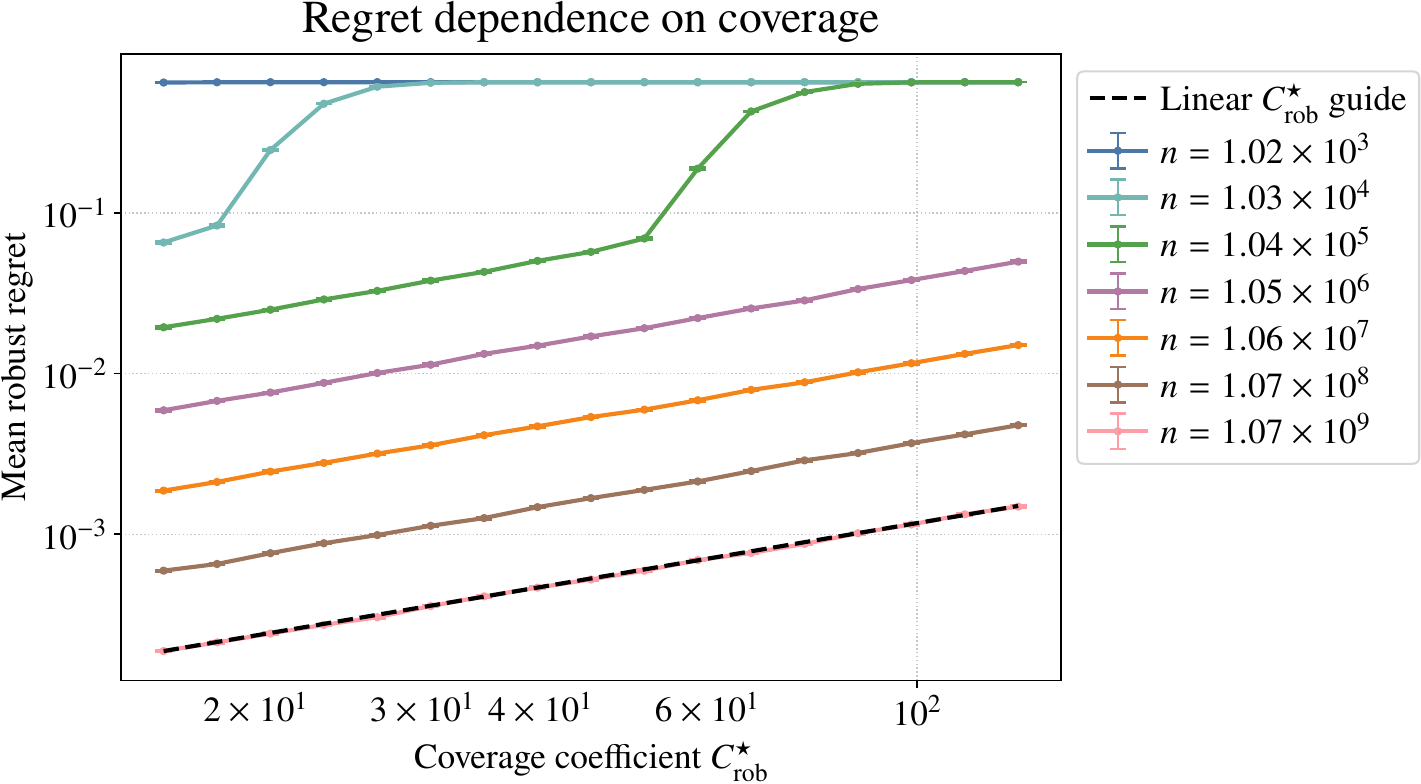}
    \caption{Mean population robust regret versus the exact coverage coefficient $C^\star_{\rm rob}$ for seven sample sizes. Error bars show $1.96$ Monte Carlo standard errors; the dashed line is a linear coverage scaling guide anchored to the largest sample size. Both axes are logarithmic.}
    \label{fig:regret-vs-c}
\end{figure}

\begin{figure}[ht]
    \centering
    \includegraphics[width=0.5\textwidth]{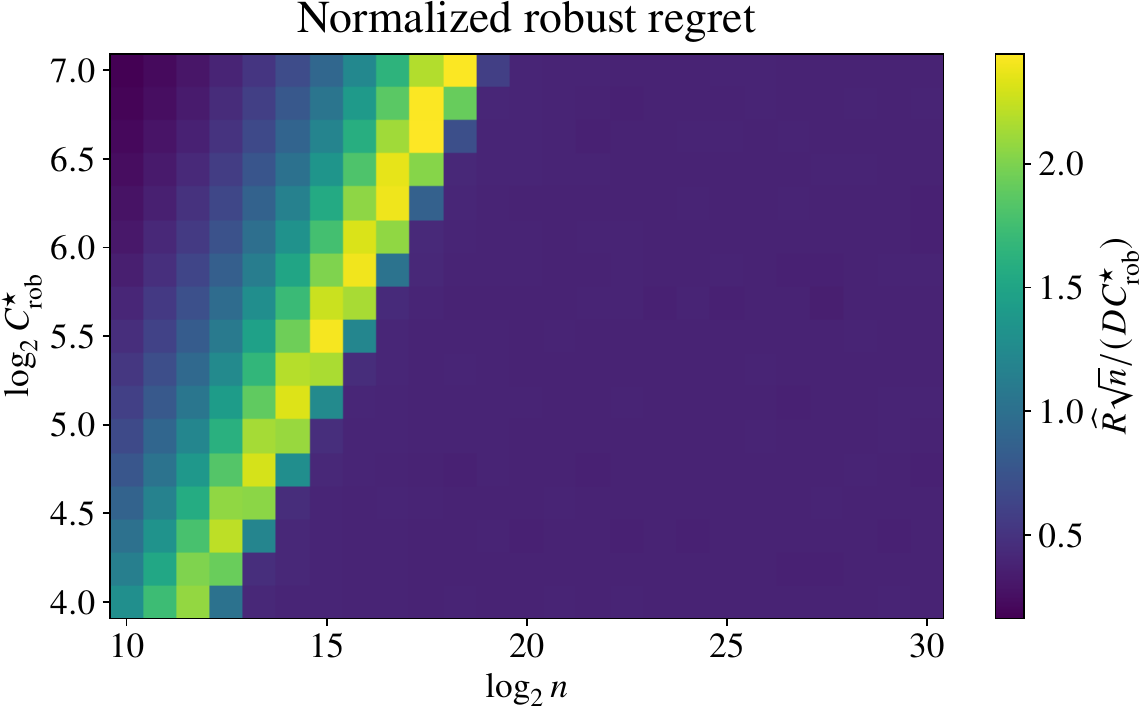}
    \caption{Normalized mean regret $\overline R\sqrt n/(D C^\star_{\rm rob})$ across all $425$ configurations of the sweep.}
    \label{fig:regret-normalized}
\end{figure}

\paragraph{Results and interpretation.}
Figures~\ref{fig:regret-vs-n} and~\ref{fig:regret-vs-c} show seven evenly spaced slices of the sweep on logarithmic axes. At $n=2^{10}$, mean regret ranges from $0.6464$ to $0.6497$ across coverage coefficients, close to the largest optimal robust value in the tested family, $q(0.65)+0.18D\approx0.6497D$. The mean fraction of contexts at which the learner chooses action $0$ ranges from $0.9719$ to $1$. The sample-size curves exhibit a plateau followed by a sharp decrease, with the transition occurring at larger sample sizes for larger coverage coefficients. Beyond this transition, the curves approach the $n^{-1/2}$ reference slope; the dependence on coverage becomes approximately linear when sample sizes are sufficiently large.

The reference lines indicate scaling only, not theoretical upper bounds. The $n^{-1/2}$ guide is anchored at the largest sample size of the middle displayed coverage slice; the linear coverage guide is anchored at the smallest coverage coefficient of the largest displayed sample size. To quantify the distinction between the full range and the large-sample behavior, we fit
\[
    \log\overline R
    =b_0+b_n\log n+b_C\log C^\star_{\rm rob}
\]
using two explicitly specified sets of configurations. A joint fit over all $425$ points gives $b_n=-0.5708$ and $b_C=1.0256$, with an $R^2$ of $0.96057$ on the logarithmic scale. These global coefficients mix the plateau, transition, and subsequent decay and are not estimates of a single large-sample power law.

When applying the lower cutoff $n\ge2^{22}$, the resulting exponents are $b_n=-0.5004$ and $b_C=1.0021$, with $R^2=0.99988$. None of these $170$ configurations selects a gap-grid endpoint. Moreover, their expected optimal-action cell counts exceed $256$, so they are well above the count range in which the confidence penalty saturates. These diagnostics support interpreting this restricted fit as a description of the observed large-sample regime; they do not define a universal threshold for such behavior.

Figure~\ref{fig:regret-normalized} provides a joint check using all configurations. Over the full range, the normalized mean regret $\overline R\sqrt n/(D C^\star_{\rm rob})$ varies from $0.1624$ to $2.4448$, making the departure from a uniform power law visible. On the restricted range $n\ge2^{22}$, it lies between $0.3585$ and $0.3910$, with mean $0.3815$ and coefficient of variation approximately $1.13\%$. Thus, in this large-sample portion of the sweep, the evaluated family exhibits approximately
\[
    \overline R\approx 0.382\,
    \frac{D C^\star_{\rm rob}}{\sqrt n}.
\]
The expanded sweep therefore shows both the finite-sample transition and subsequent behavior consistent with the predicted sample-size and coverage exponents on this family. The full curves are retained so that the restricted fit is not mistaken for a relationship holding at every sample size. These observations do not establish a universal lower bound, require this rate on every fixed instance, or validate the constants and failure probability of the theoretical high-probability bound. The logarithmic factor is constant in this experiment because $|\cX|$, $|\cA|$, and $\delta$ are fixed.

\subsection{Sensitivity to a Rare Suboptimal Action}
\label{app:experiment-rare-suboptimal}

In Appendix~\ref{app:experiment-regret-n-c}, the optimal-policy and worst-cell coverage factors coincide. We separate them by making one suboptimal action rarer while keeping the optimal-action logging probabilities fixed.

\paragraph{Instance construction.}
We retain the four uniform contexts, nine actions, reward model, and ambiguity radii of Appendix~\ref{app:experiment-regret-n-c}, with robust reward gap $\Delta=0.015$. At each context, actions $1,\ldots,7$ have logging probability $\mu=4/64^2$, action $8$ has probability $t\mu$, and the zero-reward action $0$ receives the remaining mass. Action $1$ remains optimal everywhere and each of its cells has joint probability $1/64^2$. Hence $C^\star_{\rm rob}=64$ throughout the sweep, whereas the worst-cell factor in the bound of \citet{shen2024wasserstein} is
\[
    C_{\rm worst}:=\frac{1}{\sqrt{\underline P^0\underline\pi^0}}
    =\frac{64}{\sqrt{t}}.
\]
We use $n\in\{2^{22},2^{23},2^{24}\}$ and $t\in\{1,7/8,3/4,1/2,2^{-2},2^{-4},\ldots,2^{-12}\}$. The case $t=1$ recovers the logging design of Appendix~\ref{app:experiment-regret-n-c}.

\paragraph{Paired simulation.}
For each $n$, we draw $20{,}000$ independent datasets at $t=1$. To obtain a smaller retention fraction $t$ from the previous fraction $t_{\rm prev}$, we independently retain each success and failure in the action-$8$ cells with probability $t/t_{\rm prev}$ and reassign the removed observations to action $0$ with reward zero. The resulting data have the marginal law of $n$ independent observations under the modified logger, and pairing holds the realized observations for actions $1,\ldots,7$ fixed within each trial. We use the same unclipped pessimistic learner and exact population robust-regret evaluation as in Appendix~\ref{app:experiment-regret-n-c}.

\begin{figure}[ht]
    \centering
    \includegraphics[width=\linewidth]{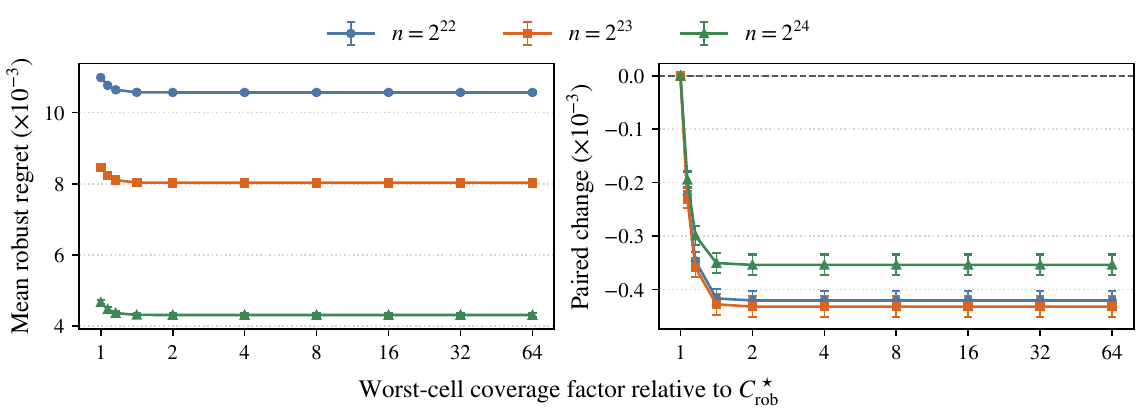}
    \caption{Robust regret as a suboptimal action becomes rarer at fixed $C^\star_{\rm rob}=64$. The left panel shows mean regret and the right panel shows its paired change from $t=1$. Error bars give approximate $95\%$ Monte Carlo intervals, computed from trial-level differences in the right panel.}
    \label{fig:rare-suboptimal-coverage}
\end{figure}

\paragraph{Results.}
Figure~\ref{fig:rare-suboptimal-coverage} shows that increasing $C_{\rm worst}$ from $64$ to $4{,}096$ reduces mean robust regret by $3.8\%$, $5.1\%$, and $7.6\%$ for $n=2^{22}$, $2^{23}$, and $2^{24}$, respectively. All three endpoint differences have paired $95\%$ Monte Carlo intervals below zero. The learner selects action $8$ in $2.8\%$--$8.1\%$ of trial--context pairs at $t=1$ and in none at $t=2^{-12}$. In this family, the larger confidence penalty makes the rare action less competitive, and regret remains controlled as worst-cell coverage deteriorates at fixed optimal-policy coverage.

\subsection{Real-Data Application: Off-Policy Learning on the Open Bandit Dataset}
\label{app:experiment-obd}

The preceding synthetic experiments compare confidence radii and regret rates with known ground truth. We complement them with off-policy evaluation on the \textbf{Open Bandit Dataset} (OBD), collected on the fashion e-commerce platform ZOZOTOWN and released with an open-source pipeline \citep{saito2020open}.

\paragraph{Dataset.}
OBD logs recommendation interactions collected under a live production system and is partitioned into three \emph{campaigns}, corresponding to different sections of the platform's catalog: \textit{all} (the union campaign, $80$ items), \textit{men} ($34$ items), and \textit{women} ($46$ items). For each campaign, OBD provides logs collected under a \textbf{uniform-random} policy. Every logged round records the item recommended, its position among the three items displayed simultaneously to the user, a binary click indicator, an anonymized user context, and the propensity score of the item actually recommended, computed directly by the logging system rather than estimated post hoc. We use the dataset's full-scale release. Each campaign--policy log is split at the row level into training, validation, and test subsets, in proportions $60\%/20\%/20\%$.

\begin{table}[ht]
    \centering
    \begin{tabular}{llrrrr}
    \hline
    Campaign & Logging policy & Rows & $|\cX|$ & $|\cA|$ & CTR (train/val/test) \\
    \hline
    men   & random & $452{,}949$ & $345$ & $34$ & $0.490\%/0.546\%/0.544\%$ \\
    women & random & $864{,}585$ & $351$ & $46$ & $0.478\%/0.490\%/0.484\%$ \\
    all   & random & $1{,}374{,}327$ & $404$ & $80$ & $0.353\%/0.349\%/0.327\%$ \\
    \hline
    \end{tabular}
    \caption{Dataset scale per campaign and logging policy: row counts, the resulting discretized context and action cardinalities, and the click-through rate (CTR) in each split.}
    \label{tab:obd-scale}
\end{table}

\paragraph{Context, action, and reward.}
We map each logged round to an i.i.d.\ draw $(x,a,y)$ from the paper's contextual-bandit model as follows.

\emph{Action.} The action $a\in\cA$ is the item recommended by the logging policy; $|\cA|\in\{34,46,80\}$ depending on the campaign (Table~\ref{tab:obd-scale}).

\emph{Context.} OBD logs four anonymized, categorical \texttt{user\_feature} fields per round, whose semantics are not disclosed by the dataset. We one-hot encode these four fields, dropping one reference level per field, and take the set of \emph{distinct} resulting rows as the finite context set $\cX$ that the paper's theory requires; this discretization is not given directly by the raw data and yields $|\cX|$ between $345$ and $404$ depending on campaign (Table~\ref{tab:obd-scale}). OBD additionally logs a \emph{position} field, since three items are displayed simultaneously per round. We do not fold position into $x$, nor otherwise model the resulting three-slot slate structure; every logged row is instead treated as an independent draw from a flat, non-slate contextual bandit.

\emph{Reward.} The reward $y\in\{0,1\}$ is the logged click indicator. The click-through rate (CTR) is low across all campaigns, between approximately $0.33\%$ and $0.55\%$ (Table~\ref{tab:obd-scale}); this low reward rate is the dominant source of statistical difficulty throughout the experiment, both for estimating $\underline q(x,a)$ from the training split and for off-policy evaluation on the test split.

\paragraph{Learner.}
We run Algorithm~\ref{alg:opl} without computing $\beta_\cX$ or using $\epsilon_\cX$. For the unrestricted policy class $\Pi=\Delta(\cA)^{|\cX|}$ used throughout this experiment, Section~\ref{ssec:pessimistic-policy-learning} shows that the per-context greedy rule $\widehat\pi(x)\in\arg\max_{a}\tilde q(x,a)$ is exactly optimal for \emph{every} $\epsilon_\cX,c_\cX$, so no linear program and no dual variable $\eta$ is required to select the policy; only the pessimistic reward table $\tilde q(x,a)$, computed independently for each cell $(x,a)\in\cX\times\cA$ from the training split, is needed.

\emph{Closed form of the robust reward.} Write $N_{x,a}$ for the number of training rounds logged with context $x$ and action $a$, and $\widehat\theta_{x,a}\in[0,1]$ for the empirical click frequency among them (the empirical CTR of cell $(x,a)$). Because the reward is Bernoulli within each cell, the reward-level dual of Lemma~\ref{lem:reward_dual} reduces to the same closed form used in Appendix~\ref{app:experiment-regret-n-c}: a nominal reward $Y\sim\operatorname{Bernoulli}(\theta)$, has robust value
\(
    q(\theta) = D\max\{\theta-\sqrt{\epsilon_\cY\theta},\,0\},
\)
and the point estimate is $\widehat q(x,a)=q(\widehat\theta_{x,a})$.

\emph{Relative reward-ambiguity radius.} We parameterize the reward-ambiguity radius relative to the training set's overall click-through rate rather than as an absolute constant, $\epsilon_\cY=\rho\cdot\widehat{\rm CTR}_{\rm train}$, for a dimensionless multiplier $\rho>0$.

\emph{Confidence radius.} With $\delta=0.1$ fixed throughout and $n_{\rm cells}=|\cX||\cA|$ the total number of context--action cells for the campaign at hand (Table~\ref{tab:obd-scale}), Proposition~\ref{prop:reward_concentration}'s confidence radius is
\[
    \beta_Y(N_{x,a};\delta)
    = D\min\left\{1,\ \sqrt{\frac{\pi}{2N_{x,a}}}+\sqrt{\frac{\log(2\,n_{\rm cells}/\delta)}{2N_{x,a}}}\right\},
\]
with the convention $\beta_Y(0;\delta)=D$. Since experiment \ref{app:experiment-q-confidence} shows that $\beta_Y$ may be too conservative, we introduce a tunable, dimensionless scale $c\ge0$ multiplying the confidence radius, giving the signed pessimistic score
\[
    \widetilde q_c(x,a) := \widehat q(x,a) - c\cdot\beta_Y(N_{x,a};\delta).
\]
The value $c=1$ recovers exactly the radius that Proposition~\ref{prop:reward_concentration} (and, in turn, Theorem~\ref{thm:coverage-rate}) is proved for; $c<1$ trades away the formal high-probability coverage guarantee for a less conservative learner.

\paragraph{Hyperparameter grid and selection.}
Two hyperparameters are swept, both on the validation split described above: the relative ambiguity radius $\rho\in\{0.01,0.03,0.1,0.3,0.6,1.0\}$ and the confidence-radius scale $c\in\{1.0,0.5,0.25,0.1,0.05,0.0\}$, giving a common $6\times6=36$-point grid for every campaign. For each $\rho$, the reward table $\widehat q(\cdot,\cdot)$ is recomputed once at $\epsilon_\cY=\rho\cdot\widehat{\rm CTR}_{\rm train}$; the inner sweep over $c$ then only rescales $\beta_Y$ and re-selects the greedy policy, reusing the same $\widehat q$ and cell counts $N_{x,a}$. Each grid point's policy is scored by a self-normalized off-policy value estimate on the validation split (defined below), and any grid point matched by fewer than $30$ validation rounds that actually resulted in a click is discarded as unreliable before selecting the best; the winning $(\rho,c)$ pair is then evaluated exactly once on the held-out test split.

\paragraph{Off-policy value estimators.}
For a policy $\pi$ and a held-out split $S=\{(x_i,a_i,y_i,\mathrm{pscore}_i)\}_{i=1}^n$, define $w_i(\pi)=\pi_{x_i}(a_i)/\mathrm{pscore}_i$ using the logged propensity $\mathrm{pscore}_i=\pi^0_{x_i}(a_i)$. For a deterministic policy, the numerator is the indicator that the logged action matches the policy's action. For stochastic POEM, it is the policy probability of the logged action. Inverse propensity scoring (IPS) and self-normalized IPS (SNIPS) are
\[
    \widehat V_{\rm IPS}(\pi)
    = \frac1n\sum_{i=1}^n w_i(\pi)y_i,
    \qquad
    \widehat V_{\rm SNIPS}(\pi)
    = \frac{\sum_{i=1}^n w_i(\pi)y_i}{\sum_{i=1}^n w_i(\pi)}.
\]
Both target the nominal value $\E_{X\sim P^0}[\sum_a\pi_X(a)\E[Y\mid X,A=a]]$. Under the logged-data model and overlap, IPS is unbiased and SNIPS is consistent. The uniform logger provides overlap by assigning positive probability to every action. We select our learner by validation SNIPS and compare all learned policies by test SNIPS. POEM uses validation IPS for selection.

\paragraph{Baselines.}
We compare Algorithm~\ref{alg:opl} with the following methods, all trained on the same campaign-specific random-log split:
\begin{itemize}
    \item \textbf{Random logging policy:} the observed test CTR provides a reference for the policy that collected the data \citep{saito2020open}.
    \item \textbf{Factored W-DRO:} the plug-in learner of \citet{shen2024wasserstein} chooses actions using empirical Wasserstein-robust rewards, without our statistical confidence penalty.
    \item \textbf{IPW Learner:} the Open Bandit Pipeline learner fits a classifier using click rewards weighted by inverse logged propensities \citep{saito2020open}.
    \item \textbf{POEM:} the stochastic linear-softmax policy of \citet{swaminathan2015counterfactual} minimizes clipped counterfactual risk with an empirical-variance penalty. We use batch optimization of that objective.
    \item \textbf{Joint KL-DRO:} the stable, self-normalized learner of \citet{si2020distributionally} optimizes worst-case value in a KL neighborhood of the joint environment distribution. We use a depth-three decision-tree policy class.
\end{itemize}

\paragraph{Results.}
Table~\ref{tab:obd-random-occupancy} reports coverage of the empirical context--action table and the hyperparameters chosen for Algorithm~\ref{alg:opl}. Table~\ref{tab:obd-results-random} compares test SNIPS. Algorithm~\ref{alg:opl} has the largest point estimate in each campaign, with a narrow margin over factored W-DRO. POEM is below the IPW Learner on women and all, but above it on men. The joint KL-DRO tree has the lowest learned-policy estimate in each campaign. These are nominal off-policy values, not estimates of worst-case deployment value or robust regret.

\begin{table}[ht]
    \centering
    \small
    \begin{tabular}{lrrrr}
    \hline
    Campaign & Occupied cells & Mean count & Selected $\rho$ & Selected $c$ \\
    \hline
    men   & $7{,}263/11{,}730$ & $37.4$ & $1.0$  & $0.25$ \\
    women & $9{,}071/16{,}146$ & $57.2$ & $0.01$ & $0.10$ \\
    all   & $18{,}049/32{,}320$ & $45.7$ & $0.03$ & $0.25$ \\
    \hline
    \end{tabular}
    \caption{Training-cell occupancy, mean count among occupied cells, and selected hyperparameters for Algorithm~\ref{alg:opl} under uniform-random logging.}
    \label{tab:obd-random-occupancy}
\end{table}

\begin{table}[ht]
    \centering
    \small
    \setlength{\tabcolsep}{4pt}
    \begin{tabular}{lrrrrrr}
    \hline
    Campaign & \shortstack{Behavior\\policy} & \shortstack{Factored\\W-DRO} & \shortstack{IPW\\Learner} & POEM & \shortstack{Joint\\KL-DRO} & Ours \\
    \hline
    men   & $0.544\%$ & $2.129\%$ & $1.087\%$ & $1.570\%$ & $0.614\%$ & $\mathbf{2.155\%}$ \\
    women & $0.483\%$ & $1.763\%$ & $1.410\%$ & $1.315\%$ & $0.936\%$ & $\mathbf{1.796\%}$ \\
    all   & $0.327\%$ & $1.734\%$ & $1.436\%$ & $1.086\%$ & $0.837\%$ & $\mathbf{1.817\%}$ \\
    \hline
    \end{tabular}
    \caption{Test-split SNIPS under uniform-random logging. POEM is stochastic; the other learned policies are deterministic. Bold marks the largest estimate in each row.}
    \label{tab:obd-results-random}
\end{table}

POEM selects $(M,\lambda)=(34,5),(46,0),(40,20)$ for men, women, and all, respectively; its test IPS values are $1.586\%$, $1.274\%$, and $1.105\%$. The selected $M$ equals the inverse random-logging propensity for men and women, so clipping is inactive there, and $\lambda=0$ removes the variance penalty for women. For joint KL-DRO, every tested KL radius yields the same tree policy on these binary-click data; the test estimates in Table~\ref{tab:obd-results-random} therefore do not identify a preferred radius.

The joint KL-DRO test SNIPS estimates rely on only $16$, $34$, and $28$ matching clicked rows for men, women, and all. For POEM, the expected numbers of logged clicked actions chosen under stochastic policy sampling are $42.3$, $47.9$, and $38.0$; these are not deterministic match counts. Such sparse click support limits precision, so the table does not establish a statistically significant ranking.

\subsection{Geographic Holdout on the International Stroke Trial}
\label{app:experiment-ist}

\paragraph{Data and split.}
We use the corrected IST release \citep{sandercock2011ist}. Contexts cross age ($<70$ or $\geq70$) and consciousness (fully conscious or drowsy/unconscious), and actions combine assigned aspirin and heparin. The six-month reward is one for \texttt{OCCODE} $3$ or $4$ (alive and independent) and zero for $1$ or $2$. Excluding $150$ unknown outcomes leaves $17{,}283$ non-test and $2{,}002$ test records. Table~\ref{tab:ist-context-action-counts} gives the pooled context--action counts.

\begin{table}[ht]
    \centering
    \small
    \setlength{\tabcolsep}{5pt}
    \begin{tabular}{@{}lrrrrrrr@{}}
    \hline
    & \multicolumn{3}{c}{No aspirin} & \multicolumn{3}{c}{Aspirin} & \\
    \cline{2-4}\cline{5-7}
    Context & None & Low & Higher & None & Low & Higher & Total (share) \\
    \hline
    $<70$, fully conscious    & $1{,}497$ & $731$     & $735$     & $1{,}469$ & $766$     & $723$     & $5{,}921$ ($30.7\%$) \\
    $<70$, drowsy/unconscious  & $291$     & $165$     & $171$     & $331$     & $149$     & $152$     & $1{,}259$ ($6.5\%$) \\
    $\geq70$, fully conscious   & $2{,}258$ & $1{,}088$ & $1{,}095$ & $2{,}183$ & $1{,}112$ & $1{,}152$ & $8{,}888$ ($46.1\%$) \\
    $\geq70$, drowsy/unconscious & $782$     & $423$     & $410$     & $833$     & $383$     & $386$     & $3{,}217$ ($16.7\%$) \\
    \hline
    Total & $4{,}828$ & $2{,}407$ & $2{,}411$ & $4{,}816$ & $2{,}410$ & $2{,}413$ & $19{,}285$ ($100\%$) \\
    \hline
    \end{tabular}
    \caption{IST context--action counts pooled across patients with observed six-month outcomes. The three columns within each aspirin group give the assigned heparin dose. Context shares are percentages of the pooled sample.}
    \label{tab:ist-context-action-counts}
\end{table}

Table~\ref{tab:ist-folds} lists the validation folds and their context radius references. Each fold trains on the non-test countries outside its validation group. For each cost and training split, the reference is the $90$th percentile of OT distances from the pooled training context distribution to the context distributions of individual training countries with at least $200$ observed outcomes. Nineteen countries appear in all training folds. The final refit covers all $24$ cells with $132$ to $2{,}063$ observations per cell.

\begin{table}[ht]
    \centering
    \small
    \setlength{\tabcolsep}{4pt}
    \begin{tabular}{@{}lrrrr@{}}
    \hline
    & & & \multicolumn{2}{c}{Context radius reference} \\
    Held-out validation countries & Train & Validation & Discrete & Age and consciousness \\
    \hline
    Belgium, Italy, Norway, Spain, UK      & $6{,}301$  & $10{,}982$ & $0.2520$ & $0.1618$ \\
    Argentina, India, New Zealand, Poland & $15{,}352$ & $1{,}931$  & $0.2238$ & $0.1426$ \\
    Portugal, Sweden, Switzerland, Turkey & $14{,}330$ & $2{,}953$  & $0.1782$ & $0.1194$ \\
    Final refit (all non-test countries)  & $17{,}283$ & ---        & $0.2217$ & $0.1464$ \\
    \hline
    \end{tabular}
    \caption{IST validation folds, record counts, and context radius references.}
    \label{tab:ist-folds}
\end{table}

\paragraph{Constrained learner and ambiguity sets.}
Let $\cA_{\rm hep}$ denote the four actions that assign either dose of heparin. We constrain every policy by
\[
    \frac{1}{4}\sum_{x\in\cX}\sum_{a\in\cA_{\rm hep}}\pi_x(a)\leq 0.30.
\]
The cap uses a uniform context distribution, so heparin shares can vary by country. With $c_{\cY}(y,y')=|y-y'|$, transporting mass from reward one to zero reduces the mean by the transport cost, down to zero. The scores in Algorithm~\ref{alg:opl} are
\[
    \underline q(x,a)
    =\max\{0,\widehat\theta_{x,a}-\epsilon_{\cY}-\beta_Y(N_{x,a};0.1)\},
\]
where $\widehat\theta_{x,a}$ is the training cell mean and $\beta_Y$ is given by~\eqref{eq:beta_Y}. We use a linear programming solver to optimize the clipped-score objective in Section~\ref{ssec:pessimistic-policy-learning}. The zero-radii control retains the confidence penalty and heparin cap.

We compare the discrete cost $\1\{x\ne x'\}$ with normalized Hamming cost, the fraction of the two context features that differ. We scale each context radius reference by $m\in\{0,0.25,0.5,1,2\}$ and pair the resulting radius with each $\epsilon_{\cY}\in\{0,0.01,0.025,0.05\}$. At $m=0$ the ambiguity sets coincide, leaving $36$ configurations.

\paragraph{Policy evaluation and selection.}
For a validation or test country with $n$ scored records, the AIPW estimate \citep{dudik2011doubly} is
\[
    \widehat V_{\rm AIPW}(\pi)
    =\frac1n\sum_{i=1}^n\left[
        \sum_{a\in\cA}\pi_{x_i}(a)\widehat\theta_{x_i,a}
        +\frac{\pi_{x_i}(a_i)}{p(a_i)}\bigl(y_i-\widehat\theta_{x_i,a_i}\bigr)
    \right].
\]
We fit the cell means on training data. For each aspirin level, $p(a)$ is $1/4$, $1/8$, or $1/8$ for no, low-dose, or higher-dose heparin, respectively. These working propensities use nominal factorial shares. The trial used minimization to balance prognostic factors \citep{ist1997trial}.

All methods use the same heparin cap and maximize equal-country validation AIPW for tuning. The IPW Learner fits multinomial logistic regression with reward divided by $p(a)$ as weights and uses the fitted probabilities as scores in a budgeted linear program. POEM optimizes IPS with a standard-error penalty of weight $\lambda$ and clips importance ratios at $8$. Stabilized joint KL-DRO searches $101$ log-spaced dual temperatures from $0.01$ to $20$ for each radius. The first candidate wins ties after rounding to $12$ decimal places. Tables~\ref{tab:ist-validation-ot} and~\ref{tab:ist-validation-baselines} report the grid results. We refit the selected policies and recompute the radius references using all scored non-test records.

\begin{table}[ht]
    \centering
    \small
    \setlength{\tabcolsep}{5pt}
    \begin{tabular}{r|rrrr|rrrr}
    \hline
    & \multicolumn{4}{c|}{Discrete} & \multicolumn{4}{c}{Age and consciousness} \\
    $m$ / $\epsilon_{\cY}$ & $0$ & $0.01$ & $0.025$ & $0.05$ & $0$ & $0.01$ & $0.025$ & $0.05$ \\
    \hline
    $0$    & $45.74$ & $45.74$ & $45.74$ & $45.74$ & --- & --- & --- & --- \\
    $0.25$ & $45.74$ & $45.74$ & $45.74$ & $45.74$ & $45.74$ & $45.74$ & $45.74$ & $45.74$ \\
    $0.5$  & $45.74$ & $45.74$ & $45.74$ & $45.74$ & $45.74$ & $45.74$ & $45.74$ & $45.74$ \\
    $1$    & $45.74$ & $45.74$ & $45.74$ & $45.74$ & $45.74$ & $45.74$ & $45.74$ & $45.74$ \\
    $2$    & $45.74$ & $45.74$ & $44.92$ & $46.05$ & $45.74$ & $45.74$ & $45.20$ & $\mathbf{46.62}$ \\
    \hline
    \end{tabular}
    \caption{Equal-country validation AIPW (\%) for every IST context cost and radius setting. Bold marks the selected configuration. The zero-radii control is the first entry of the discrete-cost grid.}
    \label{tab:ist-validation-ot}
\end{table}

\begin{table}[ht]
    \centering
    \small
    \begin{tabular}{rr|rr|rr}
    \hline
    \multicolumn{2}{c|}{IPW Learner} & \multicolumn{2}{c|}{POEM} & \multicolumn{2}{c}{Joint KL-DRO} \\
    $C$ & AIPW (\%) & $\lambda$ & AIPW (\%) & KL radius & AIPW (\%) \\
    \hline
    $0.01$ & $\mathbf{44.04}$ & $0$  & $\mathbf{44.91}$ & $0.001$ & $\mathbf{46.58}$ \\
    $0.1$  & $43.80$          & $1$  & $44.46$          & $0.003$ & $46.58$ \\
    $1$    & $43.80$          & $5$  & $44.24$          & $0.01$  & $46.58$ \\
    $10$   & $43.80$          & $10$ & $44.43$          & $0.03$  & $46.58$ \\
    ---    & ---              & $20$ & $44.52$          & $0.1$   & $46.58$ \\
    \hline
    \end{tabular}
    \caption{Equal-country validation AIPW for the tuned IST baselines. Bold marks the selected settings. All tested joint KL-DRO radii tie on validation.}
    \label{tab:ist-validation-baselines}
\end{table}

\paragraph{Held-out results and uncertainty.}
We average country-level test estimates with equal weights. In each of $1{,}000$ bootstrap draws, we resample hospitals with replacement within each country and recompute all policy estimates on the same draw. The intervals use the $5$th and $95$th percentiles of values and paired differences, conditional on the fitted policies and the four test countries.

OT-DRO has the highest country-level AIPW estimate in three countries. IPW Learner and POEM lead in Australia (Table~\ref{tab:ist-countries}). The refitted OT-DRO policy assigns heparin with probability $25\%$ under uniform context weights and $27.95\%$ to $48.71\%$ under the empirical country mixes. OT-DRO has the highest IPS estimate as well, with paired IPS intervals that include zero for IPW Learner and POEM (Table~\ref{tab:ist-ips}).

\begin{table}[ht]
    \centering
    \small
    \setlength{\tabcolsep}{4pt}
    \begin{tabular}{@{}lrrr|rrrrr@{}}
    \hline
    & & & & \multicolumn{5}{c}{AIPW (\%)} \\
    Country & Patients & Hospitals & TV & OT-DRO & Zero radii & IPW Learner & POEM & Joint KL-DRO \\
    \hline
    Australia   & $594$ & $17$ & $0.031$ & $37.57$ & $32.04$ & $42.77$ & $42.77$ & $29.35$ \\
    Austria     & $229$ & $8$  & $0.072$ & $43.18$ & $25.25$ & $16.78$ & $16.78$ & $28.38$ \\
    Czechia     & $466$ & $8$  & $0.187$ & $44.12$ & $41.25$ & $38.29$ & $38.29$ & $40.91$ \\
    Netherlands & $713$ & $10$ & $0.076$ & $48.58$ & $46.65$ & $42.46$ & $42.46$ & $44.74$ \\
    \hline
    \end{tabular}
    \caption{Test-country AIPW estimates and total-variation (TV) distances between each country's context mix and the pooled non-test mix.}
    \label{tab:ist-countries}
\end{table}

\begin{table}[!ht]
    \centering
    \small
    \begin{tabular}{lrr}
    \hline
    Policy & Test IPS (\%) & OT-DRO minus policy \\
    \hline
    OT-DRO             & $\mathbf{44.61}$ $[35.91,54.20]$ & --- \\
    Zero-radii control & $34.02$ $[28.21,40.56]$ & $10.60$ $[1.74,18.57]$ \\
    IPW Learner        & $35.97$ $[31.15,42.82]$ & $8.64$ $[-3.13,18.40]$ \\
    POEM               & $35.97$ $[31.15,42.82]$ & $8.64$ $[-3.13,18.40]$ \\
    Joint KL-DRO       & $32.86$ $[27.74,39.22]$ & $11.75$ $[2.49,19.46]$ \\
    \hline
    \end{tabular}
    \caption{Equal-country IPS estimates and paired differences with $90\%$ hospital-bootstrap intervals.}
    \label{tab:ist-ips}
\end{table}